\documentclass{ar-1col-S2O}
\usepackage[numbers]{natbib}

\usepackage{marginnote}
\usepackage{varwidth}   
\usepackage[most]{tcolorbox}
\usepackage{amsmath}
\newcounter{arbox}
\newenvironment{arbox}[1]{%
  \refstepcounter{arbox}%
  \begin{tcolorbox}[
    enhanced, breakable,
    width=\linewidth,                
    colback=white, colframe=black!70,
    boxrule=0.4pt, sharp corners,
    left=6pt,right=6pt,bottom=6pt,
    top=12pt,                        
    before skip=8pt, after skip=10pt,
    overlay={
      \node[
        anchor=south west,
        draw=black!70, fill=white,
        inner xsep=4pt, inner ysep=1.5pt,
        rounded corners=1.5pt
      ] at ([xshift=0pt,yshift=-2pt]frame.north west)
      {\bfseries Box \thearbox\ \textbar\ #1};
    },
  ]%
}{\end{tcolorbox}}

\usepackage{hyperref}
\usepackage[nameinlink,noabbrev]{cleveref} 
\crefname{arbox}{Box}{Boxes}

\usepackage{url}
\usepackage{amsmath, amsfonts, amssymb}
\usepackage{placeins}
\newcommand{\R}{\mathbb{R}}
\newcommand{\ue}{{\underline e}}
\newcommand{\ubM}{{\underline{\bm M}}}
\newcommand{\Rb}{{\bm R}}

\newcommand{\ut}{{\underline t}}

\newcommand{\mcG}{\mathcal{G}}

\usepackage{xcolor}

\def\omg{{\Omega}}

\def \bb{\mathbf{b}}

\def \ub{\mathbf{u}}

\def \xb{\mathbf{x}}

\def \xib{{\boldsymbol\xi}}

\usepackage{bm}

\newcommand{\vertii}[1]{{\left\vert\left\vert #1
    \right\vert\right\vert}}

\jname{Xxxx. Xxx. Xxx. Xxx.}
\jvol{AA}
\jyear{YYYY}
\doi{10.1146/((please add article doi))}

\firstpagenote{\textcolor{red}{Accepted for publication in the \textit{Annual Review of Biomedical Engineering} on October 2, 2025.}}

\begin{document}

\markboth{Ahmadi et al.}{Physics-Informed Machine Learning}

\title{
Physics-Informed Machine Learning in Biomedical Science and Engineering
}


\author{Nazanin Ahmadi\textsuperscript{1}, 
Qianying Cao\textsuperscript{2}, 
Jay D. Humphrey\textsuperscript{3}, 
and George Em Karniadakis\textsuperscript{2,*}
\affil{\textsuperscript{1}Center for Biomedical Engineering, Brown University, Providence, RI 02912, USA; $nazanin\_ahmadi\_daryakenari@brown.edu$}
\affil{\textsuperscript{2}Division of Applied Mathematics, Brown University, Providence, RI 02912, USA; $qianying\_cao@brown.edu$, $george\_karniadakis@brown.edu$}
\affil{\textsuperscript{3}Department of Biomedical Engineering, Yale University, New Haven, CT 06520, USA; $jay.humphrey@yale.edu$}
\affil{\textsuperscript{*}Corresponding author}}

\begin{abstract}
Physics-informed machine learning (PIML) is emerging as a potentially transformative paradigm for modeling complex biomedical systems by integrating parameterized physical laws with data-driven methods. Here, we review three main classes of PIML frameworks: physics-informed neural networks (PINNs), neural ordinary  differential equations (NODEs), and neural operators (NOs), highlighting their growing role in biomedical science and engineering. We begin with PINNs, which embed governing equations into deep learning models and have been successfully applied to biosolid and biofluid mechanics, mechanobiology, and medical imaging among other areas. We then review NODEs, which offer continuous-time modeling, especially suited to dynamic physiological systems, pharmacokinetics, and cell signaling. Finally, we discuss deep NOs as powerful tools for learning mappings between function spaces, enabling efficient simulations across multiscale and spatially heterogeneous biological domains. Throughout, we emphasize applications where physical interpretability, data scarcity, or system complexity make conventional black-box learning insufficient. We conclude by identifying open challenges and future directions for advancing PIML in biomedical science and engineering, including issues of uncertainty quantification, generalization, and integration of PIML and large language models.
\end{abstract}

\begin{keywords}
physics-informed neural networks, neural ODEs, neural operators, gray-box discovery, inverse problems.
\end{keywords}
\maketitle

\tableofcontents

\section{Introduction}
In their 2019 perspective paper, Alber et al. \cite{alber2019integrating} highlighted synergistic relationships between machine learning and multiscale modeling in biomedical science  and engineering (BSE). Machine learning can explore vast design spaces, identify correlations, construct surrogate models, and resolve ill-posed problems. Multiscale modeling, in turn, leverages underlying physics to identify causality, constrain hypothesis spaces, and elucidate biological mechanisms and chemistry.

Integration across scales and applications is naturally realized through physics-informed machine learning (PIML), a paradigm introduced in \cite{raissi2017machine,raissi2018numerical,PIML_patent} using Gaussian process regression. While such regression offers high accuracy and built-in uncertainty quantification, it lacks scalability for high-dimensional parameter spaces and is challenging to apply in strongly nonlinear settings. To overcome these limitations, PIML was reformulated using deep neural networks (NNs), introducing physics-informed neural networks (PINNs) for solving forward and inverse problems \cite{Raissi2017physicsI,raissi2017physicsII}. The unified presentation in \cite{raissi2019physics} catalyzed widespread adoption and development of PIML across scientific domains \cite{karniadakis2021physics}.

In BSE, PIML has been applied to diverse applications — from biomechanics to systems biology and quantitative pharmacology to medical imaging. PINNs integrate physical laws—typically expressed as differential equations—into the loss function alongside data fidelity terms, thereby reducing the need for expensive data assimilation while ensuring physical relevance. Physical parameters, treated analogously to NN weights, can be inferred directly during training, enabling simultaneous parameter estimation and solution discovery. Furthermore, gray-box formulations incorporate partially known physics (e.g., reaction kinetics in coagulation cascades or drug metabolism) with unknown components learned from data \cite{Ahmadi2024ai,zhang2024discovering}. The expressive power of NNs allows parameterizing governing equations with time-dependent variables, facilitating the modeling of multirate dynamics often encountered in systems biology and pharmacokinetics/pharmacodynamics (PK/PD), where 
constant-parameter models fall short.

In addition to PINNs, we also review neural ordinary differential equations (NODEs), which provide a continuous-time framework for modeling complex dynamical systems. By parameterizing rates of change of hidden states as a NN-defined vector field, NODEs learn continuous dynamics from time-series data, making them well-suited for modeling physiological processes, signaling pathways, disease progression, and PK/PD. Unlike traditional compartmental or mechanistic models, NODEs allow flexible, data-driven system identification without requiring explicit governing equations. Moreover, their compatibility with adjoint sensitivity analysis and automatic differentiation facilitates efficient training on irregularly sampled and sparse biomedical datasets.

A major advance in PIML was the introduction of deep neural operators in 2019 \cite{lu2021learning}, which map functions to functions and enable system identification from data alone. Unlike conventional NNs that operate on finite-dimensional inputs, neural operators learn solution operators of partial differential equations (PDEs), offering a higher level of abstraction ideal for complex biomedical systems. For example, in \cite{goswami2022neural}, a neural operator is trained to map mechanobiological initial conditions to aortic aneurysm progression, predicting aortic dilatation and distensibility across multiple patient-specific risk factors. Neural operators are trained offline and enable real-time inference, offering orders-of-magnitude speedup over classical solvers. Recent extensions support transfer and continual learning \cite{panos2025efficient, goswami2022deep}, accelerating deployment across patients and forming the foundation for digital twins in personalized medicine.

Figure~\ref{fig:networks} illustrates neural architectures covered in this review. In addition to classical artificial neurons, we include a recently proposed architecture motivated by the Kolmogorov–Arnold representation theorem \cite{kolmogorov1957representations}, known as the Kolmogorov–Arnold Network (KAN) \cite{liu2024kan}. KANs resemble NNs but offer improved interpretability and distinct approximation properties. We also introduce their physics-informed variant (PIKANs), which we have found particularly effective in handling sharp interfaces, stiff ODEs and noisy data, which are prevalent in biomedical modeling, especially in PK/PD systems.

\section{Physics-Informed Neural Networks (PINNs) }
PINNs were proposed in 2017 \cite{part1,part2} based on the need to seamlessly integrate data and physical models for forward and inverse problems in computational science and engineering. They are particularly effective for {\em parametric} ODEs and PDEs for which sparse data exist, even for auxiliary variables. Since their inception, there has been an explosive number of novel contributions and enhancements \cite{review1, review2}, including within the BSE community. \Cref{box:pinns} summarizes the PINN framework, where governing equations, boundary/initial conditions, and data are enforced as soft constraints in a unified loss function to learn states, parameters, and hidden dynamics.
Importantly, the forward and inverse problems use the same formulation, just the inputs/outputs differ. This simplifies modeling greatly, especially for BSE problems.\\

\begin{figure*}[h]
\includegraphics[width=\textwidth]{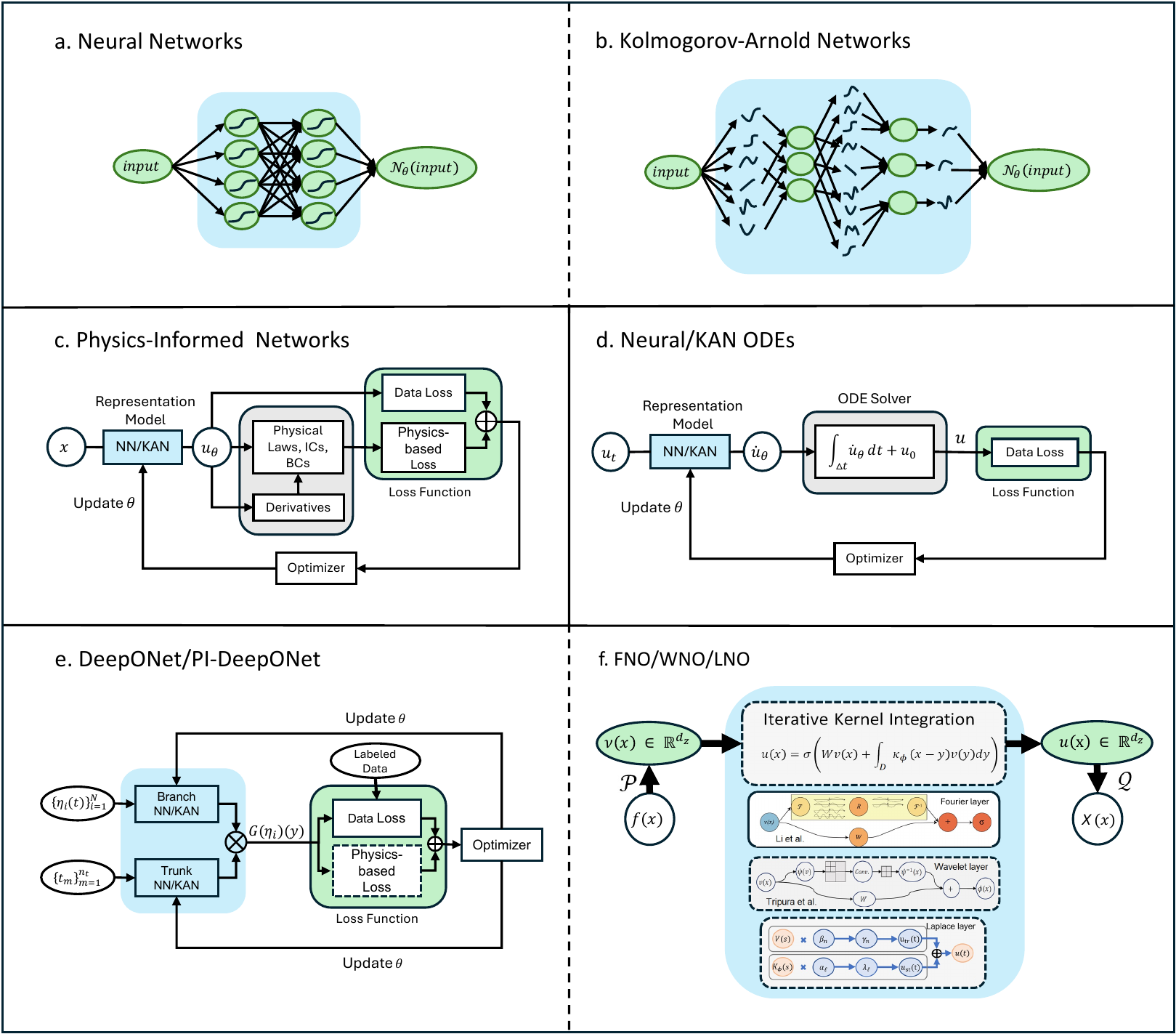}
\caption{Overview of Methods.
\textcolor{red}{(a)} Standard neural network (NN) architecture.
\textcolor{red}{(b)} Kolmogorov–Arnold Networks (KANs), which use separable inner functions $\phi_{ij}(x_j)$ and outer functions $\Phi_i$ for improved spectral representation.
\textcolor{red}{(c)} Physics-Informed Neural Networks (PINNs) and Physics-Informed Kolmogorov–Arnold Networks (PIKANs) that embed governing physical laws, initial/boundary conditions, and observed data into a unified loss function.
\textcolor{red}{(d)} NODEs or KAN-based ODEs used to model dynamic systems by solving $\dot{u} = f_\theta(u, t)$, incorporating data loss.
\textcolor{red}{(e)} DeepONet and PI-DeepONet architectures, where branch and trunk network (e.g., NN or KAN) represent operator inputs and evaluation coordinates, respectively.
\textcolor{red}{(f)} Fourier Neural Operator (FNO), Wavelet Neural Operator (WNO), and Laplace Neural Operator (LNO), which learn mappings between function spaces.
}
\label{fig:networks}
\end{figure*}

\begin{arbox}{Physics-Informed Neural Networks (PINNs)}
\phantomsection\label{box:pinns}
For illustrative purposes, a minimal PINN is outlined for the 1D reaction–diffusion equation.
We seek a surrogate \(u_\theta(x,t)\) that fits available data while driving the PDE residual to zero:
\begin{subequations}
\begin{align}
\partial_t u &= D\,\partial_{xx} u \;+\; R(u,x,t), && (x,t)\in\Omega\times(0,T],\\
u(x,0) &= u_0(x), && x\in\Omega,\\
u(x,t) &= g(x,t), && (x,t)\in\partial\Omega\times[0,T],
\end{align}
\end{subequations}
where $D>0$ is the diffusion coefficient and $R(u,x,t)$ is a (known or learned) reaction term. 
For the trainable parameters $\theta$ (the neural network weights and biases, and optionally 
unknown physical parameters such as $D$), we define the residual

\[
r_{\text{PDE}}(x,t;\theta)
= \partial_t u_\theta(x,t)\;-\;D\,\partial_{xx} u_\theta(x,t)\;-\;R\!\big(u_\theta(x,t),x,t\big).
\]

Training minimizes a weighted sum of data misfit at supervised points
\(\{(x_i,t_i,u_i)\}_{i=1}^{N_{\text{data}}}\) (IC/BC/observations) and residuals at
collocation points \(\{(x_j,t_j)\}_{j=1}^{N_r}\):
\begin{align}
\mathcal{L}(\theta)
&= w_{\text{data}}\,
    \frac{1}{N_{\text{data}}}\sum_{i=1}^{N_{\text{data}}}
    \bigl|u_\theta(x_i,t_i)-u_i\bigr|^2
 \;+\;
   w_{\text{PDE}}\,
    \frac{1}{N_r}\sum_{j=1}^{N_r}
    \bigl|r_{\text{PDE}}(x_j,t_j;\theta)\bigr|^2 .
\end{align}
The weights $ w_{\text{data}}, w_{\text{PDE}}$ can be selected manually or as hyperparameters. Derivatives are obtained by automatic differentiation. In practice, one
uses a first-order optimizer (e.g., Adam) optionally followed by L\!-\!BFGS
until convergence.

\medskip
\noindent\rule{\linewidth}{0.4pt}
\smallskip

\noindent\textbf{Representation models.}\;
Standard PINNs use multilayer perceptrons ~\cite{hornik1989multilayer} with layer map
\(z^{(l)}=\sigma\!\big(W^{(l)}z^{(l-1)}+b^{(l)}\big)\).
Enhancements include residual PINNs~\cite{wang2021understanding}, adaptive activation networks~\cite{jagtap2020adaptive}, and
Chebyshev-based Kolmogorov--Arnold Networks (KANs)~\cite{liu2024kan}. In KANs (Figure~\ref{fig:networks}b), each layer composes
learnable 1D bases \(\phi_{ij}\) with outer functions \(\Phi_i\):
\(z^{(l)}=\sum_i \Phi_i\!\left(\sum_j \phi_{ij}(z^{(l-1)}_j)\right).\)
PIKANs keep the standard PINN loss (data, constraints, PDE residuals), changing only the
representation class.

\end{arbox}

Beyond \Cref{box:pinns}, several specific choices improve PINN performance. For example, 
residual-based attention and self-adaptive weights dynamically rebalance  multiscale regions~\cite{ANAGNOSTOPOULOS2024116805,mcclenny2023self,chen2024self}. Similarly, we can mitigate well-known problems of spectral bias in NNs with feature expansions---e.g., Fourier features, random projections, or polynomial bases~\cite{wang2021eigenvector,dong2021method,cai2019multi,zhang4957859pkan}.
For multiscale or stiff PDEs, we can employ the Adam optimizer followed by L-BFGS~\cite{raissi2019physics}; second-order and hybrid methods further improve convergence~\cite{urban2025unveiling,rathore2024challenges,daryakenari2025representation}. Curriculum or sequential training reduces early overfitting to BC/IC/data terms~\cite{wang2024piratenets,cminn}.
For complex geometries, we can employ domain decomposition that aids in parallel and stable training ~\cite{jagtap2020extended,jagtap2020conservative,hu2023augmented}. For long-term integration, we can gradually expand the time windows to stabilize long-horizon learning~\cite{penwarden2023unified,cminn}.


For further details on automatic differentiation, constraint enforcement, preprocessing, and gray-box extensions, see representative works~\cite{baydin2018automatic, mao2020physics, wang2022respecting, shukla2024comprehensive, toscano2024inferring, toscano2024kkans}.

\subsection{Method Illustrated}
Among other applications, PINNs have been used widely to model biofluid mechanics, including cerebrospinal fluid (CSF) in perivascular spaces and deeper tissue in the brain. For example, an Artificial Intelligence Velocimetry (AIV) approach combines two-photon microscopy with a PINN to noninvasively reconstruct three-dimensional CSF fields in live mice ~\cite{boster2023artificial}. AIV fuses sparse single-plane particle-tracking velocimetry (PTV) with the governing equations to infer full-field velocity $\mathbf{u}(x,y,z,t)$ and pressure $p(x,y,z,t)$, enabling direct estimation of volumetric flow rate, axial pressure gradients, and wall shear stress. A follow-up study extended AIV by incorporating moving vessel-adjacent boundaries due to pulsatility and by reporting CSF quantities from the reconstructed fields ~\cite{toscano2024inferring}, including uncertainty quantification, not detailed here. The associated perivascular transport is important in waste clearance (glymphatic function), solute/drug distribution, and mechanobiological signaling; relying on plane-wise velocities alone can bias volumetric metrics and obscure pressure or shear, which are key for physiology and therapy design.
To make the reconstruction concrete, we briefly outline two practical PINN-based pipelines—surface AIV and brain-wide MR-AIV.

\textit{Surface AIV (two-photon).}
Following \cite{toscano2024inferring}, sparse PTV is acquired in one fast plane (A), while a second fast plane (B) provides vessel motion; a separate axial stack supplies the 3D geometry. PTV is \emph{phase-averaged} (ECG-synchronized) over many cycles to produce a canonical cycle, so planes acquired at different absolute times contribute coherently. The vessel width in plane B is converted to radius and differentiated in time to obtain wall velocity. Assuming axisymmetry, boundary points adjacent to the vessel are tagged as moving with the measured wall speed; the remaining boundary is treated as stationary. Collocation points for PDE residuals are generated by clustering boundary points (angular/axial), shrinking/combining subregions to fill the domain, and tiling across radial and temporal groups, yielding ordered mini-batches that stabilize training.%

\textit{Deep-brain MR-AIV.}
Complementing surface AIV, dynamic contrast-enhanced MR-AIV \cite{toscano2025mr} reconstructs brain-wide fields from time-resolved tracer concentrations without direct velocity measurements. A modular PINN enforces porous-media transport (Darcy/advection–diffusion) with separate networks for concentration (with a denoising head), pressure, and permeability; physics is imposed on the denoised signal, and optimized using time-dependent residual-based attention. The output is 3D velocity, pressure, and permeability consistent with Darcy flow, enabling maps of advection- vs. diffusion- dominated regions.

PTV provides velocities in one imaging plane and does not recover pressure; out-of- plane motion and nonuniform sampling can lead to biased volumetric estimates. A PINN augments sparse data with physical structure: the NN must satisfy governing equations and boundary conditions while fitting data. The physics-guided prior regularizes the ill-posed 3D reconstruction and enables spatial derivatives (pressure and velocity gradients) via automatic differentiation—capabilities that standard regression/interpolation or plane-wise analyses lack. Compared with classical hemodynamic simulations plus data assimilation, a PINN avoids meshing and solves forward/inverse tasks in one unified learning framework, which is advantageous with limited, noisy experimental data.

Specifically, if CSF flow is modeled using the incompressible Navier–Stokes with continuity:
\begin{align}
\partial_t \mathbf{u} + (\mathbf{u}\!\cdot\!\nabla)\mathbf{u} &= -\nabla p + \nu \nabla^2 \mathbf{u}, 
\qquad \nabla\!\cdot\!\mathbf{u}=0,
\end{align}
then in the low-Reynolds regime relevant to pial PVSs, it reduces to unsteady Stokes flow.
The boundary conditions include \emph{moving} vessel-adjacent walls inferred from axisymmetric pulsatility measured in time-series images (points adjacent to the vessel move with the vessel; other PVS boundaries are treated as stationary).

The network takes PTV tracks, the 3D geometry, and time-varying boundary motion as inputs, and predicts $(u,v,w,p)$ over space–time. Training minimizes a multi-objective loss that blends data misfit with PDE and boundary residuals at interior and boundary collocation points (see Box 1). After convergence, physiologically relevant quantities are computed from the learned fields, e.g., 
Axial pressure gradient: $\frac{\partial p}{\partial y},$
volumetric flow rate: $Q \;\approx\; \frac{A}{N}\sum_{i=1}^{N} v_i,$
and shear stress magnitude: $ \tau \;=\; \mu\,\sqrt{2\,\mathbf{D}:\mathbf{D}}, 
\qquad \mathbf{D} \;=\; \tfrac{1}{2}\big(\nabla\mathbf{u} + (\nabla\mathbf{u})^{\!\top}\big).$

\begin{figure*}[h]
\hspace*{-0.3cm}\includegraphics[width=0.9\textwidth]{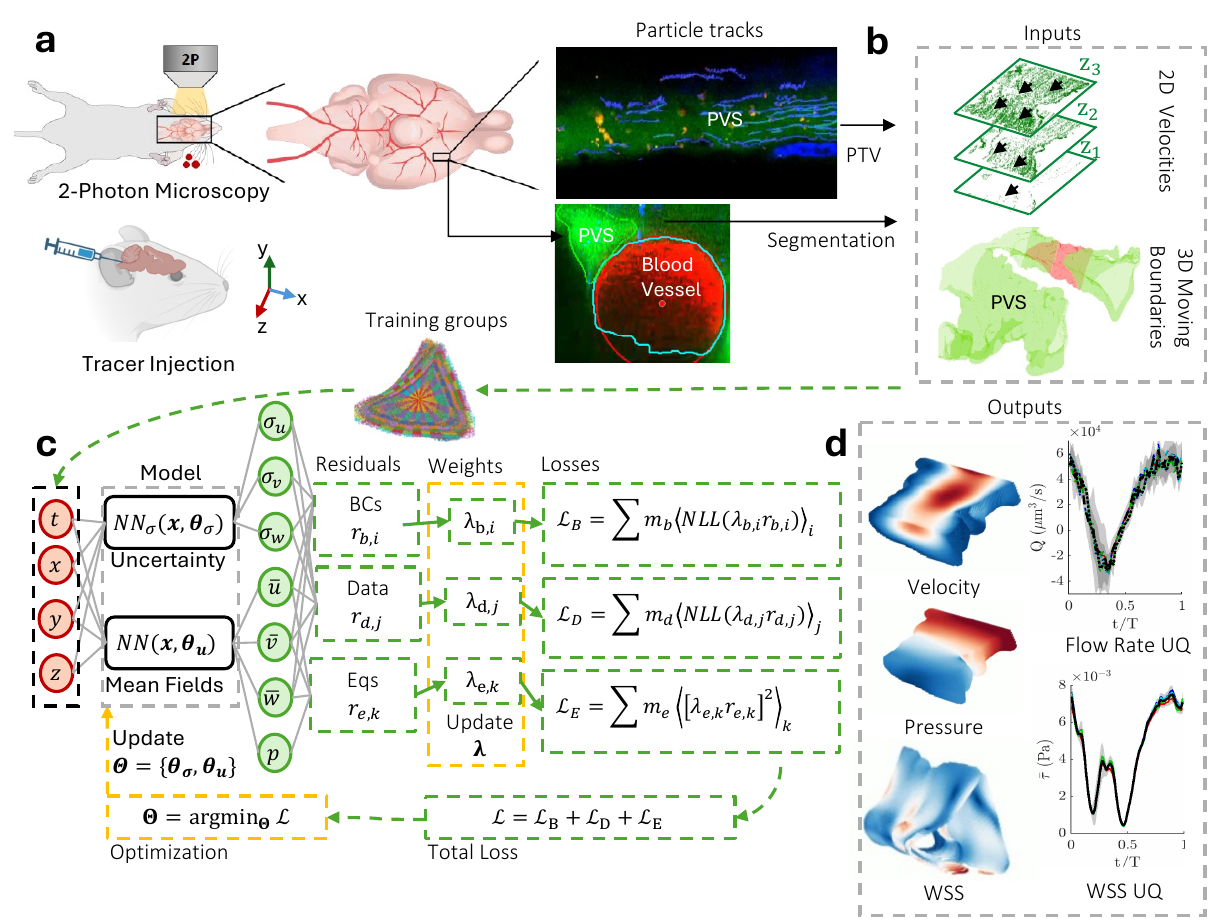}
\caption{CSF dynamics in the perivascular space of the brain of a live mouse. \textcolor{red}{(a)} Fluorescent tracers are imaged with two-photon microscopy for particle tracking and segmentation of the perivascular space (PVS). \textcolor{red}{(b)} Model inputs include sparse 2D velocities from Particle Tracking Velocimetry (PTV), the 3D domain geometry, and moving boundary conditions (MBCs) derived from segmentation. Collocation points are sampled and clustered with all data into ordered training groups. \textcolor{red}{(c)} The model assumes that the velocity is decomposed into a mean field plus Gaussian noise,
 $\boldsymbol{u} = \bar{\boldsymbol{u}} + \boldsymbol{\epsilon}_u$, where $\boldsymbol{\epsilon}_u \sim \mathcal{N}(0, \boldsymbol{\sigma}_u)$. Two NNs (with parameters $\boldsymbol{\theta}_{\bar{u}}$ and $\boldsymbol{\theta}_\sigma$) are trained to predict the mean fields ($\bar{\boldsymbol{u}}, p$) and the noise's standard deviation ($\boldsymbol{\sigma}_u$), respectively. Training minimizes a composite loss: a Negative Log-Likelihood (NLL) criterion for the noisy experimental data and boundaries and a mean-squared error for the governing PDEs.  \textcolor{red}{(d)} The trained model outputs continuous fields for velocity, pressure, and uncertainty. From these, quantities of interest like volumetric flow rate ($Q$) and wall shear stress ($\tau$) are derived over the cardiac cycle ($T$). Figure adopted from
\cite{toscano2024inferring}.} 
\label{fig:brain}
\end{figure*}

A schematic of the AIV architecture is shown in Figure~\ref{fig:brain}, adapted from \cite{toscano2024inferring}.

\subsection{Further Applications}
PINNs and other PIML NN architectures have proven useful in diverse areas of BSE, including a few highlighted below. We refer the interested reader to the original papers and references therein for details.

\subsubsection*{Systems Biology}
PINNs have increasingly been adopted for modeling dynamic behaviors in biological systems, including complex biochemical reactions ~\citep{yazdani2020systems}. Building on these earlier studies, the AI-Aristotle framework ~\citep{Ahmadi2024ai} extended the use of PINNs to hybrid gray-box discovery in both systems biology and pharmacology. Density-PINNs, which infer distributions of signal transduction times and examine how cellular variability arises in response to stimuli, address heterogeneity in signaling ~\citep{jo2024density}. Beyond ODE-based systems, Biologically-Informed Neural Networks (BINNs) learn reaction–diffusion processes from sparse in vitro data ~\citep{lagergren2020biologically}, which can uncover mechanistic insights missed by the classical models.

\subsubsection*{Soft Tissue Biomechanics}
PINNs have been used extensively to study cardiac function, including estimation of regional wall strains and biophysical properties ~\cite{caforio2024physics}, electrophysiological parameters from in silico and optical mapping data ~\cite{chiu2024characterisation}, and cardiomyofiber orientation based on solving the Eikonal equation ~\cite{ruiz2022physics}. Some extensions of these methods have been applied similarly to other soft tissues, including the brain and articular cartilage \cite{kamali2024discovering}.
PINNs have been used to construct elasticity maps of the liver based on in vivo and simulated imaging data ~\cite{ragoza2023physics}, to study perfusion of tumors by combining a diffusion-corrected extended Tofts model with dynamic contrast-enhanced MRI data ~\cite{sainz2024exploring}, to quantify skin elasticity while enforcing convexity constraints ~\cite{tac2022data}, and to infer 3D vocal fold motions from sparse 2D images constrained by modal dynamic equations ~\cite{movahhedi2023predicting}, using an LSTM-based \marginpar{\textcolor{red}{LSTM}: Long Short-Term Memory} recurrent encoder-decoder.
Hybrid frameworks inspired by PINNs have solved inverse problems in elastography to infer shear stiffness ~\cite{zhang2020physics} or more generally, via constitutive artificial neural networks (CANNs), methods for data-driven modeling and discovery of mechanical constitutive behavior of soft tissues ~\cite{linka2021constitutive, peirlinck2024automated}. Other generalized methods include a Graph NN-based surrogate emulator trained for irregular geometries using a physics-informed energy minimization approach ~\cite{dalton2023physics}, and integration of classical finite element methods with NNs to address nonlinear biosolid mechanics \cite{sacks2022neural}.


\subsubsection*{Fluid Mechanics}

PINNs have been increasingly used to simulate arterial blood flow by enforcing the incompressible Navier–Stokes equations, including via Weighted Extended PINNs and Weighted Conservative PINNs, to leverage domain decomposition and weighted loss functions to improve accuracy and parallel efficiency ~\cite{bhargava2024enhancing}. PINNs have been used to study transvalvular blood flow in laminar and turbulent regimes ~\cite{mca28020062}. PINNs can improve estimates of wall shear stress from sparse velocity measurements without known boundary conditions ~\cite{arzani2021uncovering} and recover velocity and pressure fields from dye visualizations in intracranial aneurysms ~\cite{raissi2020hfm} or predict cerebral hemodynamics in the Circle of Willis using limited clinical ultrasound and MRI data ~\cite{sarabian2022physics}. Many other examples abound.

\subsubsection*{Medical Imaging}
In image reconstruction, a noise-robust PINN-based-MREPT framework can reconstruct tissue conductivity from noisy MRI data ~\cite{inda2022physics} while another PINN can reconstruct high-resolution velocity and pressure fields from low-resolution and noisy 4D-Flow MRI data ~\cite{shone2023deep}. An optimized U-Net model, integrated with gated attention and trained using a physics-informed loss function, can similarly suppress speckle noise in ultrasound images ~\cite{hsu2024attentive}. In image registration, PINNs can improve 3D elasticity models to better define prostate motion encountered during transrectal ultrasound-guided procedures ~\cite{min2023non} as well as estimate tissue properties ~\cite{min2024biomechanics}. Similar usage of PINNs, using a hypernetwork-based method to learn physics-informed regularization for medical image registration, has been applied to 3D scans of the lung ~\cite{reithmeir2024learning, reithmeir2024data}. In image segmentation, a PINN-based multiscale feature fusion network enhanced tumor segmentation in breast ultrasound images ~\cite{ding2024pinn} while input parameterized-PINNs improved spatio-temporal resolution and reduced noise and artifacts in 4D-Flow MRI ~\cite{kalajahi2025input}.
\marginpar{\textcolor{red}{DSC}: Dynamic Susceptibility Contrast}
In field quantification, a two-step framework, sequentially combining a re-trained deep residual NN for resolution enhancement and a physics-informed image processing method, could  estimate functional relative pressure from the super-resolved data ~\cite{ferdian2023cerebrovascular}. PINNs have quantified myocardial perfusion parameters from dynamic contrast-enhanced MRI data ~\cite{van2022physics}, accurately predicted displacement, strain, and stress distributions from spinal CT data ~\cite{lin2025novel}, and solved an inverse problem in Ultrasound Computed Tomography to estimate the speed of sound in tissues ~\cite{liu2021ultrasound}. Finally, a voxel-wise PINN could  estimate the tissue residue function directly from DSC-MRI data~\cite{rotkopf2024physics}.

\subsubsection*{Systems Pharmacology}

\begin{figure*}[h]
\centering
\hspace*{-0.56cm}\includegraphics[width=\textwidth]{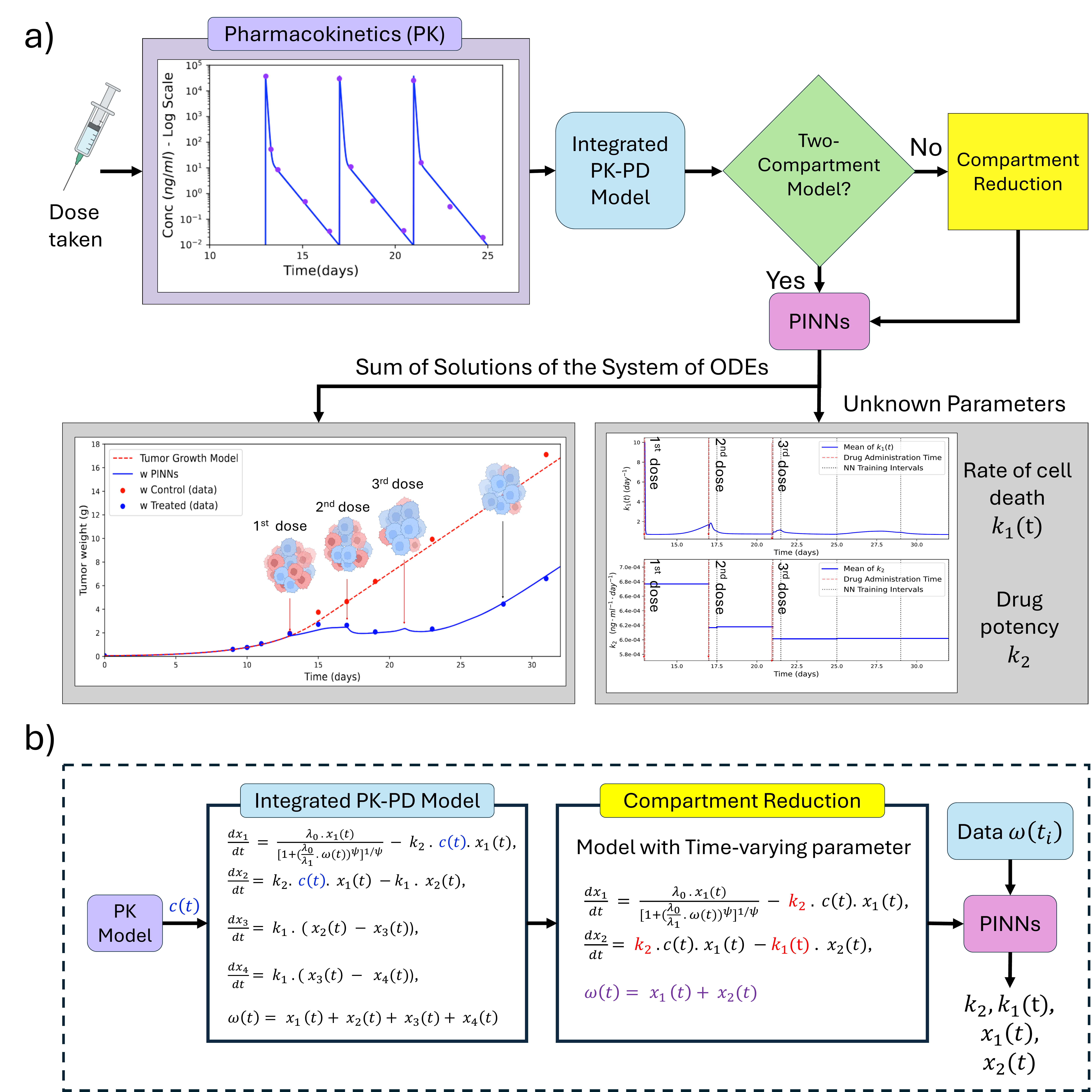}
\caption{Integrated PK–PD model within the CMINNs framework for studying tumor response under multi-dose chemotherapy. \textcolor{red}{(a)} Drug concentration over time, obtained from the pharmacokinetics model in a repeated-dose regimen, is passed into a reduced PK–PD system of two ODEs. The rate of tumor cell death, $k_{1}(t)$, is modeled as a time-varying parameter: following the first dose, $k_{1}(t)$ exhibits a sharp spike corresponding to rapid death of highly sensitive tumor cells, while subsequent doses produce progressively smaller and delayed peaks, reflecting the emergence of drug resistance and pharmacokinetic tolerance. The efficacy index $k_{2}$ is represented as a piecewise constant parameter, decreasing across treatment cycles, indicating reduced overall drug potency. Together, these dynamics capture the adaptive survival of resistant and persistent cells within the tumor microenvironment. \textcolor{red}{(b)} Workflow for reducing the high-dimensional PK–PD system to two ODEs by introducing time-varying and piecewise constant parameters, which capture changes in parameter values without adding new compartments (ODEs) to the system. Using PINNs enables simultaneous inference of unknown parameters and system trajectories. This reduction provides interpretable insights into chemotherapy resistance, tolerance, and persistence dynamics. Figure adapted from the results of Model~4 in \cite{cminn}.
.}

\label{fig:CMINN}
\end{figure*}

PINNs have emerged as powerful tools in systems pharmacology, solving differential equations that govern drug absorption, distribution, metabolism, and elimination ~\cite{goswami2023study}. PINNs can learn latent pharmacodynamic relationships ~\cite{podina2025learning}, reconstruct tissue responses under different growth assumptions such as the Verhulst and Montroll models ~\cite{rodrigues2024using}, and generalize across dose levels and patient cohorts.
The first application of PINNs in pharmacometrics was introduced in AI-Aristotle ~\cite{Ahmadi2024ai}, a framework that integrates PINNs with symbolic regression for gray-box identification. This approach enables the discovery of missing components in PK/PD models and generates interpretable, symbolic mathematical expressions from learned dynamics. Compartment Model Informed Neural Networks (CMINNs) were proposed to model nonlinear drug behavior and abrupt response changes in cancer therapies under multi-dose treatment regimens~\cite{ahmadi2024pharmacometrics}. As shown in Figure~\ref{fig:CMINN}, CMINNs reduce multi-compartment PK and PK–PD systems to simplified ODEs with time-varying or piecewise constant parameters. In PK, this allows the discovery of anomalous diffusion, trapping, and escape rates, while in PD it captures tumor drug-response dynamics such as tolerance, persistence, and resistance. Importantly, these models extend classical compartment frameworks by incorporating fractional-order derivatives, introduced in the form of fPINNs, as an equivalent representation to time-varying parameters. Further advancing gray-box system identification, recent work has compared traditional multilayer perceptron PINNs with Chebyshev-based Physics-Informed Kolmogorov–Arnold Networks~\cite{daryakenari2025representation}. The applicability of PINNs continues to expand into systems toxicology and predictive safety. For instance, HyperSBINN applied hypernetwork-enhanced PINNs for drug cardiosafety assessment to enable efficient inference from high-dimensional cardiac data ~\cite{soukarieh2024hypersbinn}. Similarly, PINNs have been used to model the electrophysiological effects of anti-arrhythmic drugs, aiding in the quantitative understanding of arrhythmia mechanisms ~\cite{chiu2024characterisation}.


\subsection{Outlook}

PINNs represent a transformative shift in computational modeling of BSE systems. Unlike traditional numerical methods that are time-consuming and often problem-specific, PINNs offer a unified and flexible framework, tackling both forward and ill-posed inverse problems. They have been widely adopted in biomechanics, systems biology, and medical imaging and have demonstrated strong potential in various applications. Despite vanilla PINNs having several limitations, modified PINNs with advanced techniques significantly enhance efficiency and accuracy. Future development should focus on: 1) enhancing generalization capabilities,
2) improving computational efficiency, and 3) automating hyperparameter-tuning and neural architecture search.
PINNs have shown promise for solving inverse problems, particularly when data are sparse but prior physical knowledge is available. Recent studies highlight several potential failure modes and inaccuracies, including ill-posedness of the inverse formulation, spectral bias toward low frequencies,   
challenges with optimization landscapes,   
and identifiability issues that can compromise parameter inference.
Some efforts to address these issues include spectral-aware training \cite{wang2020ntk}, curriculum learning strategies \cite{krishnapriyan2021characterizing}, and precision-aware optimization \cite{xu2025rethinking}. 


\section{Neural Ordinary Differential Equations (NODEs)}
NODEs have been very effective in modeling time-evolving processes in pharmacokinetics, systems biology, physiology, and medical imaging, especially when the temporal dynamics is known in part and needs to be adjusted based on data. The framework provides irregular sampling for time evolution, latent dynamics representation, and it can couple with variational inference or neural manifolds to quantify uncertainty and learn representations. \Cref{box:nodes} summarizes Neural ODEs, framing deep networks as continuous-time dynamical systems and noting common trade-offs (e.g., solver accuracy vs. speed and adjoint memory) in practice.\\

\begin{arbox}{Neural ODEs (NODEs)}
\phantomsection\label{box:nodes}
NODEs ~\cite{Neoralode} reformulate deep neural networks as
continuous-time dynamical systems. The hidden state
$\mathbf{h}(t)\in\mathbb{R}^d$ evolves according to an ODE defined by a neural
network $f_\theta$ with parameters $\theta$:
\begin{subequations}\label{eq:node}
\begin{align}
\frac{d\hat{\mathbf{h}}}{dt} &= f_\theta\!\big(\hat{\mathbf{h}}(t),\, t,\, \tau\big),
\qquad t\in[0,T], \label{eq:node_dynamics}\\
\hat{\mathbf{h}}(0) &= \mathbf{h}_0, \label{eq:node_ic}
\end{align}
\end{subequations}
where $\mathbf{h}_0$ is a known or inferred initial condition and $\tau$
collects additional physical/pharmacological parameters (e.g., clearance,
volume of distribution) that may be known or learned. The terminal state is
computed via a black-box ODE solver:
\[
\hat{\mathbf{h}}(T)
= \mathbf{h}_0 + \int_{0}^{T} f_\theta\!\big(\hat{\mathbf{h}}(t),\, t,\, \tau\big)\,dt.
\]

In gray-box or hybrid settings, the ODE function can combine known mechanics
and a neural component $\mathcal{G}_\phi$:
\[
f_\theta(\hat{\mathbf{h}}, t, \tau)
= \mathcal{F}_\tau[\hat{\mathbf{h}}] + \mathcal{G}_\phi(\hat{\mathbf{h}}, t),
\]
where $\mathcal{F}_\tau$ encodes known dynamics (e.g., compartmental PK) and $\mathcal{G}_\phi$ learns missing/unknown effects.

Given observations $\{(\mathbf{h}_i^{\text{data}}, t_i)\}_{i=1}^N$, training
minimizes the data mismatch:
\[
\mathcal{L}(\theta,\phi,\tau)
= \frac{1}{N}\sum_{i=1}^N \left\|
\hat{\mathbf{h}}(t_i) - \mathbf{h}_i^{\text{data}}
\right\|^{2},
\]
optionally augmented with regularization or physics-based constraints on $\tau$
or $\mathcal{G}_\phi$.

To solve~\eqref{eq:node}, NODEs are trained with gradient-based optimization
and adjoint sensitivity methods~\cite{Neoralode,zhuang2020adaptive}. In hybrid
models, $\tau$ and the learned dynamics $\mathcal{G}_\phi$ are inferred jointly
with $f_\theta$.
\end{arbox}

Beyond standard feedforward dynamics $f_\theta$, NODEs have evolved through extensions that improve expressivity, support uncertainty quantification, and embed structural priors. 
\subsection{Method Illustrated }
\begin{figure*}[h]
\centering
\hspace*{-0.3cm}\includegraphics[width=\textwidth]{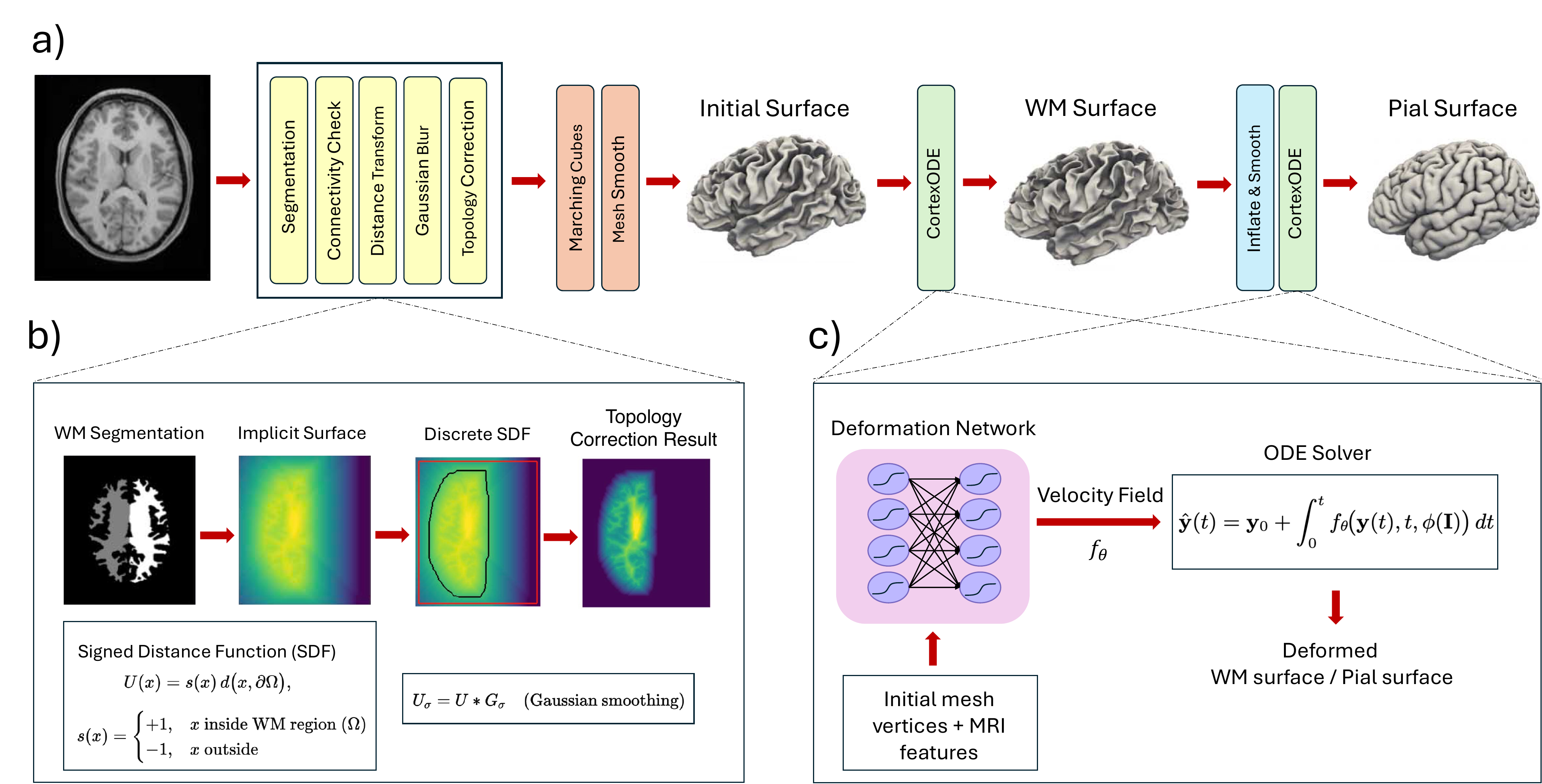}
\caption{Overview of CortexODE framework for cortical surface reconstruction.
\textcolor{red}{(a)} End-to-end pipeline: brain MRI scans are segmented to extract white matter (WM), followed by connectivity filtering, distance transform, Gaussian smoothing, and topology correction to form an implicit surface. Marching cubes and mesh smoothing generate the initial surface. Two CortexODE modules are then applied: the first deforms the initial mesh into the WM surface (optimized with Chamfer loss) and the second expands the WM surface into the pial surface (optimized with vertex-wise MSE). 
\textcolor{red}{(b)} Signed distance function (SDF) and topology correction: WM segmentation is converted into a signed distance function $U(x) = s(x)\,d(x,\partial\Omega)$, where $s(x)=+1$ inside and $-1$ outside the WM region. Gaussian smoothing yields a continuous implicit representation, and topology correction ensures genus-0 surfaces suitable for reconstruction. 
\textcolor{red}{(c)} Neural ODE core: the deformation network $f_\theta$ receives vertex coordinates and local MRI features and outputs a velocity field. Vertex trajectories are evolved via $\tfrac{dx(t)}{dt} = f_\theta(x(t),V),\; x(0)=x_0$, producing diffeomorphic deformations to WM or pial surfaces while avoiding self-intersections. Figure adopted from \cite{cortexode2023}.}
\label{fig:NODE}
\end{figure*}

Compared to other state-of-the-art learning-based frameworks,
NODEs cast deep networks as continuous-time dynamical systems, which is advantageous when the target mapping should be smooth, topology-preserving, and reversible. A representative example is cortical surface reconstruction: accurate white–matter and pial surfaces underpin morphometry, thickness mapping, and functional alignment, yet voxelwise convolution neural networks (CNNs) may introduce jaggedness or topology breaks, and classical deformable models can be slow and heavily tuned. A NODE instead learns a velocity field that continuously deforms an initial mesh into the target surfaces, naturally enforcing smooth flows and facilitating geometric regularization (see Figure~\ref{fig:NODE}; CortexODE~\cite{cortexode2023}).

\textit{Formulation.}
Let $\mathbf{y}(t) \in \R^{3V}$ denote the stacked 3D coordinates of a mesh with $V$ vertices evolving over $t\in[0,1]$. A learnable vector field $f_\theta$ drives the deformation:
\begin{subequations}\label{eq:node_surface}
\begin{align}
\frac{d\mathbf{y}}{dt} &= f_\theta\!\big(\mathbf{y}(t),\, t,\, \phi(\mathbf{I})\big), \qquad t\in[0,1], \\
\mathbf{y}(0) &= \mathbf{y}_0,
\end{align}
\end{subequations}
where $\mathbf{I}$ is the input image (e.g., T1 MRI), $\phi(\mathbf{I})$ are image features, and $\mathbf{y}_0$ is an initial mesh (atlas or coarse segmentation). The prediction is obtained by numerical integration
\[
\hat{\mathbf{y}}(t)=\mathbf{y}_0+\int_{0}^{t} f_\theta\!\big(\mathbf{y}(t),t,\phi(\mathbf{I})\big)\,dt,
\]
using a black-box ODE solver (e.g., adaptive Runge–Kutta) and trained end-to-end with the adjoint sensitivity method as in CortexODE. The objective combines data fidelity and geometric/flow regularization:
\[
\mathcal{L} = w_{\text{data}}\;\ell_{\text{surf}}\!\big(\hat{\mathbf{y}}(t),\,\mathbf{y}^{\star}\big)
+ w_{\text{geom}}\;\mathcal{R}_{\text{geom}}(\hat{\mathbf{y}}(\cdot))
+ w_{\text{flow}}\;\mathcal{R}_{\text{flow}}(f_\theta),
\]
where $\ell_{\text{surf}}$ is a surface-distance loss (e.g., point-to-surface or Chamfer), $\mathcal{R}_{\text{geom}}$ enforces mesh quality (Laplacian smoothness, edge-length regularity, normal consistency), and $\mathcal{R}_{\text{flow}}$ penalizes undesirable dynamics (e.g., excessive divergence or non-Lipschitz behavior).

\textit{Rationale for a NODE formulation.}
Relative to discrete-depth CNNs or purely variational deformation energies, a NODE provides:
(i) a continuous-time, Lipschitz-regularized flow that promotes smooth, topology-preserving deformations;
(ii) a natural vehicle for geometric priors via pathwise regularizers acting along the trajectory $\mathbf{y}(t)$ and on $f_\theta$;
(iii) adaptive integration that allocates computation where the deformation is complex; and
(iv) memory-efficient training through the adjoint method.

\textit{Outputs.}
From one forward solve, the model yields high-quality white and pial surfaces (via two coupled NODEs or conditional outputs), as well as thickness, curvature, and folding measures used in downstream neuroimaging analyses (illustrated in Figure~\ref{fig:NODE}).

\subsection{Further Applications}
NODEs and their extensions have been applied in diverse biomedical systems, from systems pharmacology to physiological modeling, oncology, brain dynamics, and medical imaging, with representative applications spanning these areas.

\subsubsection*{Systems Pharmacology}
NODEs have been applied to PK/PD modeling, showing advantages over traditional approaches in predicting new dosing regimens. Early work introduced neural PK/PD frameworks capable of learning drug concentration–response trajectories directly from clinical data, while preserving pharmacological principles and accommodating irregular sampling and heterogeneous patients~\cite{lu2021neuralODE}. These models improved temporal prediction accuracy and generalized to unseen dosing regimens. Subsequent reviews emphasized advantages in handling sparse and irregular datasets and advocated for integration into model-informed drug development~\cite{losada2024neuralODE}. Low-dimensional NODE architectures tailored for PK systems have also been developed~\cite{bram2023lowdim}.
In quantitative systems pharmacology, hybrid frameworks integrating mechanistic ODEs with NODE-driven latent dynamics enable interpretable and patient-specific forecasting, including for ICU and COVID-19 data~\cite{qian2021integrating}. 

\subsubsection*{Cardiovascular and Physiological Modeling}
In cardiovascular systems, a hybrid NODE replaced ventricular interaction terms in a lumped heart model with a neural network, later regressed into symbolic form for interpretability and robustness~\cite{grigorian2024hybrid}. A latent NODE-based digital twin encoded a 3D–0D cardiovascular model into a low-dimensional latent space, achieving 300x faster simulations of pressure–volume dynamics with 43 parameters, enabling real-time, personalized cardiac modeling~\cite{salvador2024whole}.

\subsubsection*{Oncology and Gene Regulatory Modeling}
In oncology, DN-ODE integrates NODE solvers with an encoder–decoder structure for multi-task learning and cross-phase forecasting from irregular breast cancer data~\cite{mukherjee2023dnnode}. A recent review highlights broader applications of NODEs for tumor growth and treatment modeling~\cite{metzcar2024review}. In gene regulation, Biologically Informed NODEs infer genome-wide transcriptional dynamics~\cite{hossain2023bioNODE}, while PerturbODE learns interpretable gene regulatory networks from single-cell perturbation data~\cite{lin2025interpretable}. Survival analysis has been advanced by SurvLatent-ODE, which handles competing risks and irregular sampling for time-to-event prediction~\cite{moon2022survlatent}. In single-cell analysis, SymVelo combines NODEs with dual-path architectures to estimate RNA velocity in multimodal datasets~\cite{xie2024rna}.

\subsubsection*{Medical Imaging}
Continuous-time formulations have proven especially useful in imaging, where NODEs support tasks that require smooth temporal or spatial transformations. In segmentation, U-NODE demonstrated the benefits of continuous dynamics for histological analysis~\cite{pinckaers2019neural}, later extended to brain tumor MRI and modality-specific fusion~\cite{yang2023neural}, as well as to attention-augmented hybrids for tumor detection~\cite{ru2023attention}.  
For reconstruction, NODE architectures achieve anatomically accurate results across MRI and CT, including cortical surface modeling (CortexODE)~\cite{cortexode2023} and undersampled k-space recovery (ReconODE)~\cite{chen2020mri}.  
For deformable registration, multi-scale NODE frameworks improve 3D alignment and topology preservation~\cite{xu2021multi}, while sequential NODEs capture dynamic processes such as cardiac motion and brain atrophy without requiring explicit physical priors~\cite{wu2024neural}.

\subsubsection*{Disease Progression and Brain Dynamics}
NODEs support longitudinal clinical modeling under irregular sampling. Attention-augmented frameworks have been applied to ocular disease classification~\cite{qiu2023gram}, while NODE-based models capture Alzheimer’s disease progression and impute missing data~\cite{jeong2022deep,hao2020learning}. Extensions to fMRI enable brain state classification and dynamic analysis~\cite{cai2023osr}.

\subsection{Outlook}

Despite their growing popularity across BSE, NODEs remain computationally intensive compared to alternative approaches like PINNs. This is primarily due to their reliance on black-box ODE solvers during both forward prediction and backpropagation via the adjoint method. Each iteration requires numerically integrating the neural dynamics, often with adaptive solvers, which increases training time and makes large-scale applications or high-resolution spatiotemporal modeling challenging.

In contrast to PINNs, which directly minimize physics residuals, NODEs trade off computational efficiency for smoother interpolation and greater flexibility in modeling latent or partially observed dynamics. PINNs perform best when the physics is compactly specified with clear initial and boundary conditions, but struggle with discontinuous forcings such as multi-dose drug administrations. NODEs, by learning vector fields directly from data, can incorporate such time-varying inputs naturally at inference, making them complementary to PINNs.

Recent advances such as adaptive solvers~\cite{zhuang2020adaptive}, exponential integration methods~\cite{fronk2025training}, solver-learning~\cite{chen2020mri}, and curriculum training~\cite{kidger2021efficient} aim to reduce training time and improve numerical stability. Structural variants, including hybrid NODEs~\cite{qian2021integrating, giampiccolo2024robust}, physics-constrained NODEs~\cite{kumar2025physics}, latent NODEs~\cite{rubanova2019latent}, augmented NODEs~\cite{dupont2019augmented}, and KAN-ODEs~\cite{kanode}, demonstrate how domain priors and novel architectures can enhance generalization and interpretability.

Looking ahead, key challenges include stiff dynamics, long-horizon extrapolation, and uncertainty quantification. Promising directions involve efficient Bayesian methods, solver-agnostic optimizers, and principled handling of measurement noise, alongside efforts to ensure accountability, explainability, and reproducibility for clinical or regulatory deployment.

\section{Neural Operators (NOs)}
NOs represent a higher level of abstraction than PINNs or NODEs. They are trained offline and can make predictions in almost real-time as they do not require any further training.
NOs can be considered as surrogate models, similar to reduced-order models (ROMs), but unlike ROMs which are severely under-parametrized, NOs are over-parametrized, hence, offering greater generalization ability. Specifically, NOs aim to learn mappings between function spaces:
\begin{equation}
    \mathcal{G}: f \mapsto u,
\end{equation}
where $f$ is an input function (e.g., boundary conditions or forcing terms) and $u$ is the output function, e.g., the solution of a PDE. They are particularly useful for system identification and biomedical scenarios where the physical laws are unknown. 
The first deep NO, named DeepONet, was published in 2019~\cite{lu2019deeponet} based on rigorous theoretical foundations, and later other more empirical NOs, as we present below,  were published offering different viable alternatives. \Cref{box:NO} summarizes core ideas of two main branches of NOs: DeepONet and its variants; and space-based integral neural operators. \\

\begin{arbox}{Neural operators (NO)}
\phantomsection\label{box:NO}
\noindent\textbf{Deep operator Network (DeepONet) and its variants}\;
DeepONet approximates the operator using two neural sub-networks and a latent space of dimension $p$:
\begin{equation}
\mathcal{G}_\theta(f)(y) \approx \sum_{k=1}^{p} b_k(f) \cdot t_k(y),
\end{equation}
where $\theta$ represents all trainable parameters of the model. The branch network takes the discretized input vector $[f(x_1), \dots, f(x_m)]$ and produces a latent representation $\mathbf{b}(f) = [b_1(f), \dots, b_p(f)]$, which captures global information about the function $f$. The trunk network takes the target location $y$ and outputs $\mathbf{t}(y) = [t_1(y), \dots, t_p(y)]$, encoding the spatial dependence of the operator evaluation. The output $\mathcal{G}_\theta(f)(y)$ is computed as the inner product between these two latent vectors.

Training is performed using paired data $\{(f^i, y_j, u^i(y_j))\}$ by minimizing the mean squared error between predicted and true values:
\begin{equation}
\mathcal{L}_{\text{DeepONet}} = \frac{1}{N} \sum_{i=1}^{N} \sum_{j=1}^{M} \left| \mathcal{G}_\theta(f^i)(y_j) - u^i(y_j) \right|^2,
\end{equation}
where $N$ is the number of input functions, $M$ is the number of sensor locations of the output function, and $u^i(y_j)$ denotes the true value of the function output for input $f^i$ evaluated at location $y_j$.

\medskip
\noindent\rule{\linewidth}{0.4pt}
\smallskip

\noindent\textbf{Space-based integral neural operators}\;

This type of NO is based on an iterative architecture~\cite{li2020neural}:
\begin{equation}\label{itera}
u_{t+1}(x) = \sigma\left( W u_t(x) + \int_D \kappa_{\phi}(x, y, a(x), a(y)) u_t(y) \, \nu_x(dy) \right),
\end{equation}
where $\sigma$ represents the nonlinear activation function; here, the kernel function $\kappa_{\phi}$ and the matrix $W$ are modeled as neural networks. Based on the calculation of $\kappa_{\phi}$, a series of methods were  proposed~\cite{li2020neural,li2020fourier,tripura2023wavelet,cao2024laplace}. Here, we introduce three popular methods implemented in the spectral domain.

Fourier Neural Operator (FNO) is a neural architecture designed to learn mappings between function spaces by operating in the frequency domain: 
\begin{equation}
u(x) = \mathcal{F}^{-1} \left( K_{\phi}(\omega) \cdot \left( \mathcal{F}(f(x)) \right) \right),
\end{equation}
where $K_{\phi}(\omega)$ is the Fourier transform of $\kappa_{\phi}(x)$.
The operator $\mathcal{F}(f(x))$ projects the input function into frequency space, 
and the inverse Fourier transform $\mathcal{F}^{-1}$ brings the processed signal back into the physical domain, producing the prediction $u(x)$.

Wavelet neural operator (WNO)~\cite{tripura2023wavelet} leverages discrete wavelet transforms to capture both spatial and frequency-localized features of functions:
\begin{equation}
u(x) = \mathcal{W}^{-1} \left( R_{\phi} \cdot \left( \mathcal{W}(f(x)) \right) \right),
\end{equation}
where $R_{\phi}$ is the wavelet transform of $\kappa_{\phi}(x)$. $\mathcal{W}(f(x))$ projects the input function into wavelet space, 
and the inverse wavelet transform $\mathcal{W}^{-1}$ brings the processed signal back into the physical domain and produces the prediction $u(x)$.

Laplace neural operator (LNO)~\cite{cao2024laplace} represents the output using a pole-residue formulation derived in the Laplace domain:

\begin{equation}
\label{eq:response-pol-res1}
  u(x)=\sum_{n=1}^{N}\gamma_n(\mu_n, \beta_n)\exp(\mu_n x)+\sum _{\ell=-L}^{L} \lambda_\ell(\omega_\ell, \alpha_\ell) \exp(i\omega_\ell x), 
\end{equation}
where system poles $\mu_n$ and residues $\beta_n$ are learnable parameters, used to represent $\kappa_{\phi}$, while the input poles $\omega_\ell $ and residues $\alpha_\ell$ are obtained through function decomposition methods. 

\end{arbox}

Although NOs are powerful approaches for learning solutions of general PDEs, they are data-driven only, and the learned operators are black boxes. Inspired by the idea behind PINNs, the Physics-informed NOs~\cite{wang2021learning,goswami2023physics}, which incorporate a physics-based regularization term into the loss function of the NO, have been proposed. The loss function of the Physics-informed NO is formulated as:
\begin{equation}\label{PI-Deeponet}
  \mathcal{L}(\theta) = \mathcal{L}_{\text{data}}(\theta) + \mathcal{L}_{\text{physics}}(\theta), 
\end{equation}
where $\mathcal{L}_{\text{data}}(\theta)$ is the loss associated with fitting the available measurements, and $\mathcal{L}_{\text{physics}}(\theta)$ is the physical loss that enforces that the predicted output satisfies the underlying physical laws. By adding physics into the training process, the NOs achieve greater accuracy and improved generalization, particularly in extrapolation tasks. 

\subsection{Method Illustrated}
NOs are relatively recent developments in PIML but have
already been used in diverse applications, e.g. predicting tumor growth, aneurysm growth, inferring unknown constitutive laws, cardiac dynamics, and electrophysiology. In~\cite{jafarzadeh2024peridynamic},  the Peridynamic Neural Operator (PNO) was introduced, which learns the nonlocal constitutive law in the form of a neural operator. In PNO, the neural operator architecture is designed in the form of state-based peridynamics, which is a nonlocal mechanics model based on the assumption that each material point $\xb$ is interacting within $B_\delta(\xb)$, a ball centered at $\xb$ of radius $\delta$. As such, fundamental physical laws and concepts including Galilean invariance, objectivity, and momentum balance are guaranteed.

In particular, PNO aims to learn the operator from the displacement field $\ub(\xb,t)$ to the residual with the following form:
\begin{align}
\mcG[\ub](\xb,t):=\int_{{B_\delta(\mathbf{0})}}\left(\ut[\ub,\xb,t]\langle\xib\rangle+\ut[\ub,\xb+\xib,t]\langle -\xib\rangle\right) \ubM[\ub,\xb,t]\langle\xib\rangle\;d\xib,
\end{align}
where the scalar force state $\ut$ and the nonlocal kernel function $\omega$ are parameterized as:
\begin{equation}\label{eqn:umt}
  \ut[\ub,\xb,t]\langle{\xib}\rangle:= \sigma^{NN}(\omega(\xb,\xib), \vartheta(\xb,t), \ue[\ub,\xb,t]\langle\xib\rangle, |\xib|;\theta_t)\text{ ,}
\end{equation}
\begin{equation}\label{eqn:omega_aniso}
  \omega(\xib) :=\omega^{NN}(\xb,\xib;\theta_\omega)\text{.}
\end{equation}
Here, $\sigma^{NN}$ and $\omega^{NN}$ are scalar-valued functions that take the form of a shallow MLP with learnable parameters $\theta_t$ and $\theta_\omega$, respectively.  $\ue[\ub,\xb,t]\langle\xib\rangle$, $\vartheta(\xb,t)$, and $\ubM[\ub,\xb,t]\langle\xib\rangle$ are physical quantities representing the length changes of the bond, the nonlocal dilatation, and the direction of the deformed bond, respectively. The kernel function $\omega(\xb,\xib)$ characterizes the nonlocal interactions between each material point $\xb$ and its neighborhood point $\xb+\xib$, and it contains physics interpretation about material properties. For instance, in homogeneous and isotropic materials one can set $\omega(\xib) :=\omega^{NN}(|\xib|;\theta_\omega)$, and for heterogeneous and fiber-reinforced materials such as cardiovascular tissues one can encode $\omega$ as
\begin{equation}\label{eqn:omega_heter}
  \omega(\xb, \xib) :=\omega^{NN}(\Rb(-\alpha(\xb; \theta_\alpha)) \xib;\theta_\omega)\text{ , where }\Rb(\alpha) :=
\begin{bmatrix}
    \mathrm{cos} \alpha & -\mathrm{sin} \alpha \\
    \mathrm{sin} \alpha & \mathrm{cos} \alpha
\end{bmatrix}
\end{equation}
is the rotation matrix~\cite{jafarzadeh2025heterogeneous}. By learning the point-wise rotation angle $\alpha(\xb; \theta_\alpha)$ simultaneously with the PNO model, the fiber orientation field can also be inferred from data.

Given $S_{tr}$ numbers of samples composed by measured displacement field $\ub^s(\xb,t^k)$ and the applied body force $\bb^s(\xb,t^k)$, boundary condition $\ub^s_{BC}(\xb,t^k)$, $s=1,\cdots,S_{tr}$, $k=1,\cdots,N$, PNO is trained by minimizing the data error:
\begin{equation}
\theta_t^*,\theta_\omega^*,\theta_\alpha^*=\underset{\theta_t,\theta_\omega,\theta_\alpha}{\text{argmin}}\;loss_u,
\end{equation}
with
\begin{equation}\label{eqn: caseII_loss}
   loss_u := \frac{1}{N\,S_{tr}}\sum_{k = 1}^{N}\sum_{s = 1}^{S_{tr}}\dfrac{\vertii{\mcG^{-1}[\bb^s,\ub^s_{BC}](\cdot,t^k)-\ub^s(\cdot,t^k)}_{l^2(\omg)}}{\vertii{\ub^s(\cdot,t^k)}_{l^2(\omg)}}\text{ ,}
\end{equation}
where $\mcG^{-1}[\bb^s,\ub^s_{BC}]$ denotes the numerical solution of $\rho\ddot{\ub}^s=\mcG[\ub]+\bb^s$ subject to the boundary condition $\ub^s_{BC}(\xb,t)$, using an iterative nonlinear solver. The outcomes are two-fold. First, a nonlocal constitutive equation
\begin{equation}
\rho\ddot{\ub}(\xb,t)=\mcG[\ub](\xb,t)+\bb(\xb,t)
\end{equation}
is obtained, which can be employed in downstream material deformation predictions subject to new and unseen loadings $\bb^{test}(\xb,t)$, boundary conditions $\ub_{BC}^{test}(\xb,t)$, and even computational domains. Second, physical discoveries on material properties are obtained, providing new knowledge about the specimen. For instance, $\alpha(\xb; \theta_\alpha)$ recovers the collagen fiber orientation field from displacement measurements, enabling new understanding on the interrelationship between tissue mechanics and collagen microstructure.

Because of its hard-coded physical laws, PNO features physical interpretability and improved generalizability when learning from scarce and incomplete data. Hence, it serves as a perfect candidate for material modeling from experimental measurements. As an exemplar validation of this capability,  ~\cite{jafarzadeh2025heterogeneous} applied PNO on the modeling of tissues by considering digital image correlation (DIC) measurements from biaxial mechanical testing.  As shown in Figure 4 (a-b), a representative tricuspid valve anterior leaflet (TVAL) specimen was obtained from a porcine heart and mounted to a biaxial testing device. Then, seven protocols of displacement-controlled testing were performed, targeting various biaxial stresses: $P_{11}:P_{22}=1:1$, $1:0.75$, $1:0.5$, $1:0.25$, $0.75:1$, $0.5:1$, $0.25:1$. As a result, displacement and loading fields at $1398$ time instances were collected, among which $120$ instances were reserved for the purpose of training and validation, and $1098$ instances were for testing. The test errors (average relative l2-norm) of heterogeneous PNO are reported in Figure 4 (c), in comparison with a PNO model without heterogeneity (denoted as HomoPNO), and the classical Fung's models with fitted parameters. One can see that learning the nonlocal constitutive law together with material heterogeneity substantially reduces the test errors by $75\%$. This finding validates the downstream prediction capability of PNOs. In Figure 4 (d-e), the interpretability capability is demonstrated. The model is capable of predicting the Piola-Kirchhoff stress field, which is not provided by DIC measurements. Additionally, the predicted collagen fiber orientation field also matches well with the collagen fiber quantification from polarized spatial frequency domain imaging (pSFDI) device.

\begin{figure*}[h]
\includegraphics[width=\textwidth, trim = 0cm 19cm 0cm 2cm, clip]{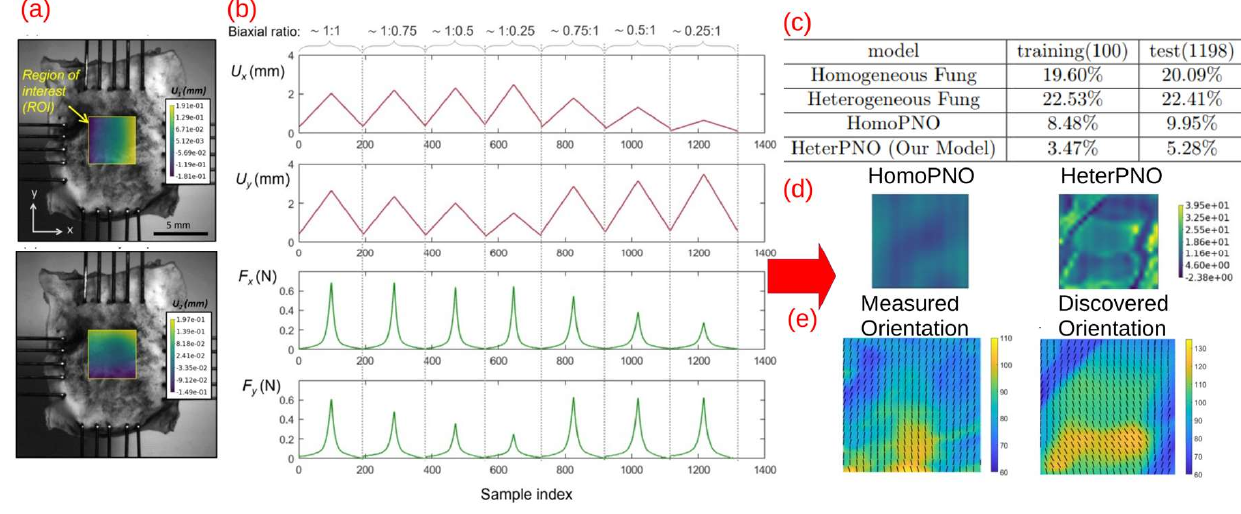}\\
\caption{Biaxial mechanical testing data from a representative porcine tricuspid valve anterior leaflet (TVAL) tissue (a-b), and results from constitutive nonlocal operator models (PNO) trained on this dataset (c-e). \textcolor{red}{(a)} Demonstration of displacement field measurements via digital image correlation (DIC). \textcolor{red}{(b)} Displacements and forces of the tines in the biaxial test of all 1398 samples, collected from seven protocols (loading-unloading cycles). \textcolor{red}{(c)} Averaged relative l2-norm error of different models’ prediction for the displacement field given boundary conditions. Here, homoPNO denotes the constitutive nonlocal operator model with homogeneous collagen fiber orientation, and heterPNO is the constitutive model with heterogeneous fiber orientation. \textcolor{red}{(d)} Predicted first Piola-Kirchhoff stress field (in kPa) computed from DIC-measured displacements using the heterPNO model. \textcolor{red}{(e)} Prediction of collagen fiber orientation field by the heterPNO model versus the experimentally detected orientation.
Figure adopted from \cite{jafarzadeh2025heterogeneous}.} 
\label{fig:PNO}
\end{figure*}

\subsection{Further Applications}
NOs have shown effectiveness across a range of applications in BSE, some of which are highlighted below.

\subsubsection*{Tissue mechanics and biomedical imaging}
NOs provide a new data-driven alternative for extracting biomechanical information directly from experimental or simulated measurement. In soft tissue mechanics, NOs are trained to infer full-field deformations from sparse boundary displacement measurements or denoise measured displacement fields in ultrasound elastography without requiring explicit PDE models~\cite{you2022physics,zhu2024neural}. They are also used to solve the inverse problem, namely estimating the tissue elasticity from displacement fields~\cite{tripura2023wavelet}. Similar approaches have been applied in cardiovascular mechanics, where operators infer aortic pressure waveforms from an inlet flow rate waveform and predict mechanobiological insult profiles based on aortic dilatation and distensibility~\cite{hong2024estimating,goswami2022neural}. Applications have also been adopted for advanced medical imaging tasks, where NOs predict focused ultrasound pressure distributions and generate aeration maps directly from raw ultrasound signals~\cite{kumar2024convolutional,wang2025ultrasound}, as well as improve image registration~\cite{drozdov2024fnoreg}. At the genetic scale, NOs are used to link mutations in structural proteins to tissue-level dysfunction~\cite{zhang2022g2varphinet}. This work offers a powerful tool for understanding genetic contributions to biomechanical pathology.

\subsubsection*{Biomedical dynamics and disease progression}
NOs have also been increasingly applied to capture biomedical dynamics and disease progression. Examples include predicting full-field dermal displacement over time, tracking the evolution of intramural damage fields in the aortic wall, and simulating the full spatiotemporal propagation of photoacoustic waves from an initial pressure distribution~\cite{husanovic2024deep,yin2022simulating,guan2023fourier}. Building on these foundations, more advanced NOs such as TGM-ONets and LFLDNets have been introduced to predict long-term tumor growth, to model cardiac electrophysiology and vascular hemodynamics, and to learn pharmacokinetics in data-scarce settings~\cite{chumburidze2025efficient,salvador2024liquid,hu2024state}. These extended methods enhance the capability of vanilla NOs by incorporating mechanisms such as attention, mixture-of-experts, or Mamba. Continued innovation along these lines is expected to yield increasingly specialized operator frameworks for addressing complex problems in biomedical engineering.

\subsection{Outlook}
NOs have been successfully adapted to address diverse problems in BSE, offering real-time performance, strong generalization, and the ability to encode physics. Studies that model time evolution illustrate the growing importance of dynamic predictions in biomedical applications. 
Future work should focus on the following points. First, incorporating physical information and prior knowledge into NOs can help to address the challenge of data scarcity. Second, predicting accurately outside the training distribution is a key current challenge for NOs. That is, future research should focus on how to leverage advanced techniques to improve the generalization and extrapolation ability of NOs. Third, foundation models in biomechanics are important to enable fast and generalizable predictions. We should establish foundation models by training on diverse biomedical datasets, which can then be fine-tuned for specific pathologies and tasks using minimal task-specific data and reduced training time. Fourth, biological structures in biomechanics, such as tissues and organs, often exhibit highly irregular geometries and heterogeneous properties. We should  leverage graph-based NOs to model irregular structures in biology, such as tumors or complex vascular networks.




\section{Needs and Directions in PIML}
Reductionist approaches have led to remarkable advances in BSE. Nevertheless, there is an ever-increasing need for integrative approaches, ones that can describe and predict the inherent complexity of intracellular signaling networks, cell-cell and cell-matrix interactions, and ultimately organ-level physiology and pathophysiology. Given the increasingly larger data-bases available, particularly via omics approaches, scientific machine learning promises to contribute significantly by synthesizing data to yield new understanding. Yet, all biological processes must respect fundamental biochemical and biophysical rules, including mass, momentum, and energy balance as well as the second law of thermodynamics. PIML thus offers great promise in enabling the integration and prediction needed to advance BSE and, as reviewed herein, myriad approaches apply to diverse applications. Much has been learned, yet much remains to be accomplished. Here, we highlight a few areas that merit attention.

Similarly to classical methods, predictions based on PIML should incorporate total uncertainty quantification (UQ). In biomedical applications, uncertainty arises from multiple intertwined sources that reflect both the complexity of living systems and the limitations of data acquisition. Epistemic uncertainty arises from limited or incomplete data, especially common in personalized medicine and rare disease modeling, where patient-specific measurements may be sparse or missing. Aleatoric uncertainty reflects intrinsic biological variability and sensor noise, such as fluctuations in heart rate, gene expression, or blood pressure readings, which cannot be reduced by collecting more data. Structural uncertainty emerges when mathematical models only partially capture physiological mechanisms, as seen in gray-box models that approximate but do not fully encode biochemical or biomechanical dynamics. Input uncertainty is prevalent due to variability in patient parameters, such as age, weight, or drug metabolism rates, while output uncertainty arises from measurement noise and missing modalities in clinical diagnostics. Robust UQ is thus critical in PINNs, NODEs, and NOs, especially as they are increasingly applied in data-limited and safety-critical biomedical domains. Work in this direction is emerging using Bayesian or ensemble approaches for NODEs \cite{d2020bayesian, giampiccolo2024robust} and PINNs and neural operators \cite{zouuq}.

Another critical need in BSE modeling is the efficient use of multifidelity and multimodality data. In practical scenarios, researchers often have access to a limited set of high-fidelity data, including experimental results or high-resolution finite element simulations. While these data may be accurate, they are usually too sparse to train robust and generalizable machine learning models. On the other hand, low-fidelity data generated either via lower resolution measurements or through simplified models, such as reduced-order models or low-resolution simulations, are more abundant and computationally accessible. Although such data lack accuracy, they can capture essential system behaviors and provide a cost-effective way to pre-train or guide machine learning models. Multifidelity training is thus a practical and powerful strategy for different PIML approaches, such as PINNs and neural operators \cite{meng2020composite,lu2022multifidelity}. Future research should continue to explore and develop hybrid learning frameworks that strategically combine data across multiple fidelity levels to make full use of all available information to build robust and generalizable models, which can pave the way for data-efficient learning in complex applications.\\

\marginpar{\textcolor{red}{PET}: Positron Emission Tomography}
\marginpar{\textcolor{red}{SPECT}: Single Photon Emission Computed Tomography}
Future research should continue to explore and develop hybrid learning frameworks that strategically combine data across multiple fidelity levels. Such approaches can make full use of all available information sources to build robust and generalizable models, which can pave the way for data-efficient learning in complex system biology applications.

In addition to multifidelity strategies, 
multimodality learning is another promising direction for advancing modeling and analysis in biological systems. Recent advances make it possible to simultaneously acquire multiple high-dimensional data modalities, such as medical imaging, time-series physiological signals, and omics profiles. Even within medical imaging alone, over ten major modalities exist, such as MRI, CT, PET, SPECT, ultrasound, and X-ray. Each modality provides a unique view of the underlying biological system, together capturing complementary features that cannot be fully represented by any single data type. Effectively fusing multimodality data can improve model robustness, enhance predictive performance, and increase generalization. Recent studies have begun to explore multimodality learning frameworks that integrate information from diverse modalities, for example, combining PET and CT data ~\cite{song2013optimal}, or integrating transcriptomics, electrophysiology, and DNA data ~\cite{tang2023explainable}. Notwithstanding advantages, integration of multimodality data introduces new challenges. First, different modalities often differ in terms of dimensionality, scale, and noise, which makes harmonization nontrivial. Second, alignment across modalities is frequently imperfect either due to differences in resolution or sampling context, which may cause inconsistency or error propagation in downstream tasks. Developing robust fusion strategies that can accommodate modality-specific characteristics and exploit shared information remains an open and critical area of research.

Beyond these technical challenges and opportunities, there is the possibility of a larger paradigm shift in computational modeling of BSE systems. For example, the emerging use of foundation models will prompt new developments that will have a similar effect as ImageNet had on imaging and segmentation. To this end, neural operators can be used as the backbone of foundation models (e.g., specific function) while an ensemble of pretrained neural operators can be used at the organ level. Similarly, we anticipate the integration of PINNs with large language models (LLMs) and Agents due to their complementary strengths. The latter excel at code generation, reasoning, workflow automation, and knowledge synthesis, but they do not ``understand" PDEs or physics constraints directly. A potential integration pathway is to use LLMs as orchestrators of PINNs, where Agents could set up PINN architectures automatically, e.g., choose collocation points, boundary conditions, optimizers, hyperparameters. Moreover, LLMs could translate natural language biomedical problems (“simulate oxygen diffusion in tumors with hypoxia-induced reactions”) into PINN formulations. The opposite is also possible and promising: Agents can call PINNs as ``modules” inside broader workflows. For example, an AI lab assistant planning drug delivery experiments might use an LLM for reasoning but query a PINN for drug diffusion in tissue under specific boundary conditions. In another instance, PINNs could provide structured physics-constrained embeddings of BSE systems that LLMs could incorporate into their reasoning (e.g., coupling PINN-learned latent variables with text-based knowledge about physiology). There are already new paradigms for autonomous laboratories and accelerated scientific discovery, e.g. \cite{swanson2024virtual,essam2025robin}, and inclusion of PINNs in such frameworks will make them more quantitative, scalable, and robust.


\section*{DISCLOSURE STATEMENT}
GEK has a small holding in the companies PhinyxAI and PredictiveIQ.
The other authors are not aware of any affiliations, memberships, funding, or financial holdings that
might be perceived as affecting the objectivity of this review. 

\section*{ACKNOWLEDGMENTS}
This work was supported, in part, by a grant from the NIH (R01 HL168473) to GEK (as well as R. Assi and C. Bellini). Also, GEK would like to acknowledge support from NIH grant R01AT012312, and a subaward (00000450) under NIH grant R01AT012312.

\begin{verbatim}

\end{verbatim}
\bibliographystyle{ar-style3}  
\bibliography{main}           

\begin{thebibliography}{152}
\expandafter\ifx\csname natexlab\endcsname\relax\def\natexlab#1{#1}\fi

\bibitem{alber2019integrating}
Alber M, Buganza~Tepole A, Cannon WR, De S, Dura-Bernal S, et~al. 2019.
Integrating machine learning and multiscale modeling—perspectives, challenges, and opportunities in the biological, biomedical, and behavioral sciences.
\textit{NPJ digital medicine} 2(1):115

\bibitem{raissi2017machine}
Raissi M, Perdikaris P, Karniadakis GE. 2017{\natexlab{a}}.
Machine learning of linear differential equations using gaussian processes.
\textit{Journal of Computational Physics} 348:683--693

\bibitem{raissi2018numerical}
Raissi M, Perdikaris P, Karniadakis GE. 2018.
Numerical gaussian processes for time-dependent and nonlinear partial differential equations.
\textit{SIAM Journal on Scientific Computing} 40(1):A172--A198

\bibitem{PIML_patent}
Raissi M, Perdikaris P, Em KG. 2021.
Physics informed learning machines.
\textit{US Patent 10,963,540}

\bibitem{Raissi2017physicsI}
Raissi M, Perdikaris P, Karniadakis GE. 2017{\natexlab{b}}.
Physics informed deep learning (part i): Data-driven solutions of nonlinear partial differential equations.
\textit{arXiv preprint arXiv:1711.10561}

\bibitem{raissi2017physicsII}
Raissi M, Perdikaris P, Karniadakis GE. 2017{\natexlab{c}}.
Physics informed deep learning (part ii): Data-driven discovery of nonlinear partial differential equations.
\textit{arXiv preprint arXiv:1711.10566}

\bibitem{raissi2019physics}
Raissi M, Perdikaris P, Karniadakis GE. 2019.
Physics-informed neural networks: A deep learning framework for solving forward and inverse problems involving nonlinear partial differential equations.
\textit{Journal of Computational Physics} 378:686--707

\bibitem{karniadakis2021physics}
Karniadakis GE, Kevrekidis IG, Lu L, Perdikaris P, Wang S, Yang L. 2021.
Physics-informed machine learning.
\textit{Nature Reviews Physics} 3(6):422--440

\bibitem{Ahmadi2024ai}
Daryakenari NA, Florio MD, Shukla K, Karniadakis GE. 2024.
{AI-Aristotle}: A physics-informed framework for systems biology gray-box identification.
\textit{PLOS Computational Biology} 20(3):e1011916

\bibitem{zhang2024discovering}
Zhang Z, Zou Z, Kuhl E, Karniadakis GE. 2024{\natexlab{a}}.
Discovering a reaction--diffusion model for alzheimer’s disease by combining pinns with symbolic regression.
\textit{Computer Methods in Applied Mechanics and Engineering} 419:116647

\bibitem{lu2021learning}
Lu L, Jin P, Pang G, Zhang Z, Karniadakis GE. 2021{\natexlab{a}}.
Learning nonlinear operators via {DeepONet} based on the universal approximation theorem of operators.
\textit{Nature machine intelligence} 3(3):218--229

\bibitem{goswami2022neural}
Goswami S, Li DS, Rego BV, Latorre M, Humphrey JD, Karniadakis GE. 2022{\natexlab{a}}.
Neural operator learning of heterogeneous mechanobiological insults contributing to aortic aneurysms.
\textit{Journal of the Royal Society Interface} 19(193):20220410

\bibitem{panos2025efficient}
Panos A, Aljundi R, Reino DO, Turner RE. 2025.
Efficient few-shot continual learning in vision-language models.
\textit{arXiv preprint arXiv:2502.04098}

\bibitem{goswami2022deep}
Goswami S, Kontolati K, Shields MD, Karniadakis GE. 2022{\natexlab{b}}.
Deep transfer operator learning for partial differential equations under conditional shift.
\textit{Nature Machine Intelligence} 4(12):1155--1164

\bibitem{kolmogorov1957representations}
Kolmogorov AN. 1957.
\textit{On the representations of continuous functions of many variables by superposition of continuous functions of one variable and addition}.
In \textit{Dokl. Akad. Nauk USSR}, vol. 114, pp.  953--956

\bibitem{liu2024kan}
Liu Z, Wang Y, Vaidya S, Ruehle F, Halverson J, et~al. 2024.
{KAN: Kolmogorov-Arnold Networks}.
\textit{arXiv preprint arXiv:2404.19756}

\bibitem{part1}
Raissi M, Perdikaris P, Karniadakis GE. 2017{\natexlab{d}}.
Physics informed deep learning (part i): Data-driven solutions of nonlinear partial differential equations.
\textit{arXiv preprint arXiv:1711.10561}

\bibitem{part2}
Raissi M, Perdikaris P, Karniadakis GE. 2017{\natexlab{e}}.
Physics informed deep learning (part ii): Data-driven discovery of nonlinear partial differential equations.
\textit{arXiv preprint arXiv:1711.10566}

\bibitem{review1}
Toscano JD, Oommen V, Varghese AJ, Zou Z, Daryakenari NA, et~al. 2025{\natexlab{a}}.
From {PINNs} to {PIKANs}: Recent advances in physics-informed machine learning.
\textit{Machine Learning for Computational Science and Engineering} 1(1):1--43

\bibitem{review2}
Raissi M, Perdikaris P, Ahmadi N, Karniadakis GE. 2024.
Physics-informed neural networks and extensions.
\textit{arXiv preprint arXiv:2408.16806}

\bibitem{hornik1989multilayer}
Hornik K, Stinchcombe M, White H. 1989.
{Multilayer feedforward networks are universal approximators}.
\textit{Neural Networks} 2(5):359--366

\bibitem{wang2021understanding}
Wang S, Teng Y, Perdikaris P. 2021.
{Understanding and mitigating gradient flow pathologies in physics-informed neural networks}.
\textit{SIAM Journal on Scientific Computing} 43(5):A3055--A3081

\bibitem{jagtap2020adaptive}
Jagtap AD, Kawaguchi K, Karniadakis GE. 2020.
{{Adaptive activation functions accelerate convergence in deep and physics-informed neural networks}}.
\textit{Journal of Computational Physics} 404:109136

\bibitem{ANAGNOSTOPOULOS2024116805}
Anagnostopoulos SJ, Toscano JD, Stergiopulos N, Karniadakis GE. 2024.
Residual-based attention in physics-informed neural networks.
\textit{Computer Methods in Applied Mechanics and Engineering} 421:116805

\bibitem{mcclenny2023self}
McClenny LD, Braga-Neto UM. 2023.
{Self-adaptive physics-informed neural networks}.
\textit{Journal of Computational Physics} 474:111722

\bibitem{chen2024self}
Chen W, Howard AA, Stinis P. 2024.
{Self-adaptive weights based on balanced residual decay rate for physics-informed neural networks and deep operator networks}.
\textit{arXiv preprint arXiv:2407.01613}

\bibitem{wang2021eigenvector}
Wang S, Wang H, Perdikaris P. 2021{\natexlab{a}}.
{{On the eigenvector bias of Fourier feature networks: From regression to solving multi-scale PDEs with physics-informed neural networks}}.
\textit{Computer Methods in Applied Mechanics and Engineering} 384:113938

\bibitem{dong2021method}
Dong S, Ni N. 2021.
{{A method for representing periodic functions and enforcing exactly periodic boundary conditions with deep neural networks}}.
\textit{Journal of Computational Physics} 435:110242

\bibitem{cai2019multi}
Cai W, Xu ZQJ. 2019.
{Multi-scale deep neural networks for solving high dimensional PDEs}.
\textit{arXiv preprint arXiv:1910.11710}

\bibitem{zhang4957859pkan}
Zhang Z, Shen T, Zhang Y, Zhang W, Wang Q. 2024{\natexlab{b}}.
{AL-PKAN: A Hybrid GRU-KAN Network with Augmented Lagrangian Function for Solving PDEs}.
\textit{Available at SSRN 4957859}

\bibitem{urban2025unveiling}
Urb{\'a}n JF, Stefanou P, Pons JA. 2025.
Unveiling the optimization process of {Physics Informed Neural Networks}: How accurate and competitive can pinns be?
\textit{Journal of Computational Physics} 523:113656

\bibitem{rathore2024challenges}
Rathore P, Lei W, Frangella Z, Lu L, Udell M. 2024.
{Challenges in training {PINNs}}: A loss landscape perspective.
\textit{arXiv preprint arXiv:2402.01868}

\bibitem{daryakenari2025representation}
Daryakenari NA, Shukla K, Karniadakis GE. 2025.
Representation meets optimization: Training {PINNs} and {PIKANs} for gray-box discovery in systems pharmacology.
\textit{arXiv preprint arXiv:2504.07379}

\bibitem{wang2024piratenets}
Wang S, Li B, Chen Y, Perdikaris P. 2024.
{Piratenets: Physics-informed deep learning with residual adaptive networks}.
\textit{Journal of Machine Learning Research} 25(402):1--51

\bibitem{cminn}
{Ahmadi Daryakenari} N, Wang S, Karniadakis G. 2025.
{CMINNs}: Compartment model informed neural networks — unlocking drug dynamics.
\textit{Computers in Biology and Medicine} 184:109392

\bibitem{jagtap2020extended}
Jagtap AD, Karniadakis GE. 2020.
{Extended physics-informed neural networks ({XPINNs}}): A generalized space-time domain decomposition based deep learning framework for nonlinear partial differential equations.
\textit{Communications in Computational Physics} 28(5)

\bibitem{jagtap2020conservative}
Jagtap AD, Kharazmi E, Karniadakis GE. 2020.
{Conservative physics-informed neural networks on discrete domains for conservation laws: Applications to forward and inverse problems}.
\textit{Computer Methods in Applied Mechanics and Engineering} 365:113028

\bibitem{hu2023augmented}
Hu Z, Jagtap AD, Karniadakis GE, Kawaguchi K. 2023.
{Augmented Physics-Informed Neural Networks ({APINNs}}): A gating network-based soft domain decomposition methodology.
\textit{Engineering Applications of Artificial Intelligence} 126:107183

\bibitem{penwarden2023unified}
Penwarden M, Jagtap AD, Zhe S, Karniadakis GE, Kirby RM. 2023.
{A unified scalable framework for causal sweeping strategies for physics-informed neural networks ({PINNs}}) and their temporal decompositions.
\textit{Journal of Computational Physics} 493:112464

\bibitem{baydin2018automatic}
Baydin AG, Pearlmutter BA, Radul AA, Siskind JM. 2018.
{Automatic differentiation in machine learning: a survey}.
\textit{Journal of Machine Learning Research} 18(153):1--43

\bibitem{mao2020physics}
Mao Z, Jagtap AD, Karniadakis GE. 2020.
{Physics-informed neural networks for high-speed flows}.
\textit{Computer Methods in Applied Mechanics and Engineering} 360:112789

\bibitem{wang2022respecting}
Wang S, Sankaran S, Perdikaris P. 2022.
{Respecting causality is all you need for training physics-informed neural networks}.
\textit{arXiv preprint arXiv:2203.07404}

\bibitem{shukla2024comprehensive}
Shukla K, Toscano JD, Wang Z, Zou Z, Karniadakis GE. 2024.
{A comprehensive and {FAIR}} comparison between {MLP} and {KAN} representations for differential equations and operator networks.
\textit{Computer Methods in Applied Mechanics and Engineering} 431:117290

\bibitem{toscano2024inferring}
Toscano JD, Wu C, Ladr{\'o}n-de Guevara A, Du T, Nedergaard M, et~al. 2024.
Inferring in vivo murine cerebrospinal fluid flow using artificial intelligence velocimetry with moving boundaries and uncertainty quantification.
\textit{Interface Focus} 14(6):20240030

\bibitem{toscano2024kkans}
Toscano JD, Wang LL, Karniadakis GE. 2024.
{KKANs: Kurkova-Kolmogorov-Arnold Networks and Their Learning Dynamics}.
\textit{arXiv preprint arXiv:2412.16738}

\bibitem{boster2023artificial}
Boster KA, Cai S, Ladr{\'o}n-de Guevara A, Sun J, Zheng X, et~al. 2023.
Artificial intelligence velocimetry reveals in vivo flow rates, pressure gradients, and shear stresses in murine perivascular flows.
\textit{Proceedings of the National Academy of Sciences} 120(14):e2217744120

\bibitem{toscano2025mr}
Toscano JD, Guo Y, Wang Z, Mori Y, Vaezi M, et~al. 2025{\natexlab{b}}.
{MR-AIV} reveals in-vivo brain-wide fluid flow with physics-informed {AI}.
\textit{bioRxiv} :2025--07

\bibitem{yazdani2020systems}
Yazdani A, Lu L, Raissi M, Karniadakis GE. 2020.
Systems biology informed deep learning for inferring parameters and hidden dynamics.
\textit{PLOS Computational Biology} 16(11):e1007575

\bibitem{jo2024density}
Jo H, Hong H, Hwang HJ, Chang W, Kim JK. 2024.
{Density physics-informed neural networks reveal sources of cell heterogeneity in signal transduction}.
\textit{Patterns} 5(2)

\bibitem{lagergren2020biologically}
Lagergren JH, Nardini JT, Baker RE, Simpson MJ, Flores KB. 2020.
Biologically-informed neural networks guide mechanistic modeling from sparse experimental data.
\textit{PLoS computational biology} 16(12):e1008462

\bibitem{caforio2024physics}
Caforio F, Regazzoni F, Pagani S, Karabelas E, Augustin C, et~al. 2024.
Physics-informed neural network estimation of material properties in soft tissue nonlinear biomechanical models.
\textit{Computational Mechanics} :1--27

\bibitem{chiu2024characterisation}
Chiu CE, Pinto AL, Chowdhury RA, Christensen K, Varela M. 2024.
\textit{Characterisation of Anti-Arrhythmic Drug Effects on Cardiac Electrophysiology Using Physics-Informed Neural Networks}.
In \textit{2024 IEEE International Symposium on Biomedical Imaging (ISBI)}, pp.  1--5. IEEE

\bibitem{ruiz2022physics}
Ruiz~Herrera C, Grandits T, Plank G, Perdikaris P, Sahli~Costabal F, Pezzuto S. 2022.
Physics-informed neural networks to learn cardiac fiber orientation from multiple electroanatomical maps.
\textit{Engineering with Computers} 38(5):3957--3973

\bibitem{kamali2024discovering}
Kamali A, Laksari K. 2024.
Discovering 3d hidden elasticity in isotropic and transversely isotropic materials with physics-informed unets.
\textit{Acta Biomaterialia} 184:254--263

\bibitem{ragoza2023physics}
Ragoza M, Batmanghelich K. 2023.
\textit{Physics-informed neural networks for tissue elasticity reconstruction in magnetic resonance elastography}.
In \textit{International Conference on Medical Image Computing and Computer-Assisted Intervention}, pp.  333--343. Springer

\bibitem{sainz2024exploring}
Sainz-DeMena D, P{\'e}rez M, Garc{\'\i}a-Aznar JM. 2024.
Exploring the potential of physics-informed neural networks to extract vascularization data from dce-mri in the presence of diffusion.
\textit{Medical Engineering \& Physics} 123:104092

\bibitem{tac2022data}
Tac V, Sree VD, Rausch MK, Tepole AB. 2022.
Data-driven modeling of the mechanical behavior of anisotropic soft biological tissue.
\textit{Engineering with Computers} 38(5):4167--4182

\bibitem{movahhedi2023predicting}
Movahhedi M, Liu XY, Geng B, Elemans C, Xue Q, et~al. 2023.
Predicting 3d soft tissue dynamics from 2d imaging using physics informed neural networks.
\textit{Communications Biology} 6(1):541

\bibitem{zhang2020physics}
Zhang E, Yin M, Karniadakis GE. 2020.
Physics-informed neural networks for nonhomogeneous material identification in elasticity imaging.
\textit{arXiv preprint arXiv:2009.04525}

\bibitem{linka2021constitutive}
Linka K, Hillg{\"a}rtner M, Abdolazizi KP, Aydin RC, Itskov M, Cyron CJ. 2021.
Constitutive artificial neural networks: A fast and general approach to predictive data-driven constitutive modeling by deep learning.
\textit{Journal of Computational Physics} 429:110010

\bibitem{peirlinck2024automated}
Peirlinck M, Linka K, Hurtado JA, Kuhl E. 2024.
On automated model discovery and a universal material subroutine for hyperelastic materials.
\textit{Computer Methods in Applied Mechanics and Engineering} 418:116534

\bibitem{dalton2023physics}
Dalton D, Husmeier D, Gao H. 2023.
Physics-informed graph neural network emulation of soft-tissue mechanics.
\textit{Computer Methods in Applied Mechanics and Engineering} 417:116351

\bibitem{sacks2022neural}
Sacks MS, Motiwale S, Goodbrake C, Zhang W. 2022.
Neural network approaches for soft biological tissue and organ simulations.
\textit{Journal of biomechanical engineering} 144(12):121010

\bibitem{bhargava2024enhancing}
Bhargava S, Chamakuri N. 2024.
Enhancing arterial blood flow simulations through physics-informed neural networks.
\textit{arXiv preprint arXiv:2404.16347}

\bibitem{mca28020062}
Du~Toit JF, Laubscher R. 2023.
Evaluation of physics-informed neural network solution accuracy and efficiency for modeling aortic transvalvular blood flow.
\textit{Mathematical and Computational Applications} 28(2)

\bibitem{arzani2021uncovering}
Arzani A, Wang JX, D'Souza RM. 2021.
Uncovering near-wall blood flow from sparse data with physics-informed neural networks.
\textit{Physics of Fluids} 33(7)

\bibitem{raissi2020hfm}
Raissi M, Yazdani A, Karniadakis GE. 2020.
Hidden fluid mechanics: Learning velocity and pressure fields from flow visualizations.
\textit{Science} 367(6481):1026--1030

\bibitem{sarabian2022physics}
Sarabian M, Babaee H, Laksari K. 2022.
Physics-informed neural networks for brain hemodynamic predictions using medical imaging.
\textit{IEEE transactions on medical imaging} 41(9):2285--2303

\bibitem{inda2022physics}
Inda AJG, Huang SY, {\.I}mamo{\u{g}}lu N, Qin R, Yang T, et~al. 2022.
Physics informed neural networks ({PINN}) for low snr magnetic resonance electrical properties tomography (mrept).
\textit{Diagnostics} 12(11):2627

\bibitem{shone2023deep}
Shone F, Ravikumar N, Lassila T, MacRaild M, Wang Y, et~al. 2023.
\textit{Deep physics-informed super-resolution of cardiac 4D-flow MRI}.
In \textit{International Conference on Information Processing in Medical Imaging}, pp.  511--522. Springer

\bibitem{hsu2024attentive}
Hsu WT, Agbodike O, Chen J. 2024.
\textit{Attentive U-Net with Physics-Informed Loss for Noise Suppression in Medical Ultrasound Images}.
In \textit{2024 10th International Conference on Applied System Innovation (ICASI)}, pp.  409--411. IEEE

\bibitem{min2023non}
Min Z, Baum ZM, Saeed SU, Emberton M, Barratt DC, et~al. 2023.
\textit{Non-rigid medical image registration using physics-informed neural networks}.
In \textit{International Conference on Information Processing in Medical Imaging}, pp.  601--613. Springer

\bibitem{min2024biomechanics}
Min Z, Baum ZM, Saeed SU, Emberton M, Barratt DC, et~al. 2024.
\textit{Biomechanics-informed Non-rigid Medical Image Registration and its Inverse Material Property Estimation with Linear and Nonlinear Elasticity}.
In \textit{International Conference on Medical Image Computing and Computer-Assisted Intervention}, pp.  564--574. Springer

\bibitem{reithmeir2024learning}
Reithmeir A, Schnabel JA, Zimmer VA. 2024.
\textit{Learning physics-inspired regularization for medical image registration with hypernetworks}.
In \textit{Medical Imaging 2024: Image Processing}, vol. 12926, pp.  625--635. SPIE

\bibitem{reithmeir2024data}
Reithmeir A, Felsner L, Braren R, Schnabel JA, Zimmer VA. 2024.
\textit{Data-Driven Tissue-and Subject-Specific Elastic Regularization for Medical Image Registration}.
In \textit{International Conference on Medical Image Computing and Computer-Assisted Intervention}, pp.  575--585. Springer

\bibitem{ding2024pinn}
Ding J, Zhu B, Wang W, Zhang S, Zhua D, Liua Z. 2024.
{PINN-EMFNet: PINN-based} and enhanced multi-scale feature fusion network for breast ultrasound images segmentation.
\textit{arXiv preprint arXiv:2412.16937}

\bibitem{kalajahi2025input}
Kalajahi AP, Csala H, Mamun ZB, Yadav S, Amili O, et~al. 2025.
Input parameterized physics informed neural networks for de noising, super-resolution, and imaging artifact mitigation in time resolved three dimensional phase-contrast magnetic resonance imaging.
\textit{Engineering Applications of Artificial Intelligence} 150:110600

\bibitem{ferdian2023cerebrovascular}
Ferdian E, Marlevi D, Schollenberger J, Aristova M, Edelman ER, et~al. 2023.
Cerebrovascular super-resolution 4d flow mri--sequential combination of resolution enhancement by deep learning and physics-informed image processing to non-invasively quantify intracranial velocity, flow, and relative pressure.
\textit{Medical Image Analysis} 88:102831

\bibitem{van2022physics}
van Herten RL, Chiribiri A, Breeuwer M, Veta M, Scannell CM. 2022.
Physics-informed neural networks for myocardial perfusion mri quantification.
\textit{Medical Image Analysis} 78:102399

\bibitem{lin2025novel}
Lin R, Zhang J. 2025.
A novel approach to biomechanical modeling: Ct image weight initialization and physics informed neural networks.
\textit{Biomedical Signal Processing and Control} 109:107939

\bibitem{liu2021ultrasound}
Liu X, Almekkawy M. 2021.
\textit{Ultrasound computed tomography using physical-informed neural network}.
In \textit{2021 IEEE International Ultrasonics Symposium (IUS)}, pp.  1--4. IEEE

\bibitem{rotkopf2024physics}
Rotkopf LT, Ziener CH, von Knebel-Doeberitz N, Wolf SD, Hohmann A, et~al. 2024.
A physics-informed deep learning framework for dynamic susceptibility contrast perfusion mri.
\textit{Medical physics} 51(12):9031--9040

\bibitem{goswami2023study}
Goswami K, Sharma A, Pruthi M, Gupta R. 2023{\natexlab{a}}.
Study of drug assimilation in human system using physics informed neural networks.
\textit{International Journal of Information Technology} 15(1):315--324

\bibitem{podina2025learning}
Podina L, Ghodsi A, Kohandel M. 2025.
Learning chemotherapy drug action via universal physics-informed neural networks.
\textit{Pharmaceutical Research} :1--20

\bibitem{rodrigues2024using}
Rodrigues JA. 2024.
Using physics-informed neural networks {(PINNs)} for tumor cell growth modeling.
\textit{Mathematics} 12(8):1195

\bibitem{ahmadi2024pharmacometrics}
Ahmadi N, Wang S, Karniadakis G. 2024.
Pharmacometrics modeling via physics-informed neural networks: Integrating time-variant absorption rates and fractional calculus for enhancing prediction accuracy.
\textit{arXiv preprint arXiv:2412.21076}

\bibitem{soukarieh2024hypersbinn}
Soukarieh I, Hessler G, Minoux H, Mohr M, Schmidt F, et~al. 2024.
{HyperSBINN}: A hypernetwork-enhanced systems biology-informed neural network for efficient drug cardiosafety assessment.
\textit{arXiv preprint arXiv:2408.14266}

\bibitem{wang2020ntk}
Wang S, Yu X, Perdikaris P. 2022.
When and why {PINNs} fail to train: A neural tangent kernel perspective.
\textit{Journal of Computational Physics} 449:110768

\bibitem{krishnapriyan2021characterizing}
Krishnapriyan A, Gholami A, Zhe S, Kirby R, Mahoney MW. 2021.
Characterizing possible failure modes in physics-informed neural networks.
\textit{Advances in neural information processing systems} 34:26548--26560

\bibitem{xu2025rethinking}
Xu C, Liu D, Nassereldine A, Xiong J. 2025.
Fp64 is all you need: Rethinking failure modes in physics-informed neural networks.
\textit{arXiv preprint arXiv:2505.10949}

\bibitem{Neoralode}
Chen RTQ, Rubanova Y, Bettencourt J, Duvenaud DK. 2018.
\textit{Neural Ordinary Differential Equations}.
In \textit{Advances in Neural Information Processing Systems}, ed. S~Bengio, H~Wallach, H~Larochelle, K~Grauman, N~Cesa-Bianchi, R~Garnett, vol.~31. Curran Associates, Inc.

\bibitem{zhuang2020adaptive}
Zhuang J, Duvenaud D. 2020.
Adaptive checkpoint adjoint method for gradient estimation in neural odes.
\textit{ICML}

\bibitem{cortexode2023}
Ma Q, Li L, Robinson EC, Kainz B, Rueckert D, Alansary A. 2023.
{CortexODE: Learning Cortical Surface Reconstruction by Neural ODEs}.
\textit{IEEE Transactions on Medical Imaging} 42(2):430--443

\bibitem{lu2021neuralODE}
Lu J, Deng K, Zhang X, Liu G, Guan Y. 2021{\natexlab{b}}.
{Neural-ODE} for pharmacokinetics modeling and its advantage to alternative machine learning models in predicting new dosing regimens.
\textit{iScience} 24:102804

\bibitem{losada2024neuralODE}
Losada IB, Terranova N. 2024.
Bridging pharmacology and neural networks: A deep dive into neural ordinary differential equations.
\textit{CPT: Pharmacometrics \& Systems Pharmacology} 13(8):1289--1296

\bibitem{bram2023lowdim}
Br{\"a}m S, Frey N, Wicha SG. 2023.
Low-dimensional neural odes and their application in pharmacokinetics.
\textit{Journal of Pharmacokinetics and Pharmacodynamics} 50:625--637

\bibitem{qian2021integrating}
Qian Z, Zame W, Fleuren L, Elbers P, van~der Schaar M. 2021.
Integrating expert odes into neural odes: pharmacology and disease progression.
\textit{Advances in Neural Information Processing Systems} 34:11364--11383

\bibitem{grigorian2024hybrid}
Grigorian G, George SV, Lishak S, Shipley RJ, Arridge S. 2024.
A hybrid neural ordinary differential equation model of the cardiovascular system.
\textit{Journal of the Royal Society Interface} 21(212):20230710

\bibitem{salvador2024whole}
Salvador M, Strocchi M, Regazzoni F, Augustin CM, Dede’ L, et~al. 2024.
Whole-heart electromechanical simulations using latent neural ordinary differential equations.
\textit{NPJ Digital Medicine} 7(1):90

\bibitem{mukherjee2023dnnode}
Mukherjee S, et~al. 2023.
{DN-ODE: Data-driven neural-ODE modeling for breast cancer tumor dynamics and progression-free survivals}.
\textit{Computers in Biology and Medicine} 161:106918

\bibitem{metzcar2024review}
Metzcar J, Jutzeler CR, Macklin P, K{\"o}hn-Luque A, Br{\"u}ningk SC. 2024.
A review of mechanistic learning in mathematical oncology.
\textit{Frontiers in Immunology} 15:1363144

\bibitem{hossain2023bioNODE}
Hossain I, Fanfani V, Fischer J, Quackenbush J, Burkholz R. 2024.
Biologically informed neuralodes for genome-wide regulatory dynamics.
\textit{Genome Biology} 25(1):127

\bibitem{lin2025interpretable}
Lin Z, Chang S, Zweig A, Kang M, Azizi E, Knowles DA. 2025.
Interpretable neural odes for gene regulatory network discovery under perturbations.
\textit{arXiv preprint arXiv:2501.02409}

\bibitem{moon2022survlatent}
Moon I, Groha S, Gusev A. 2022.
Survlatent ode: A neural ode based time-to-event model with competing risks for longitudinal data improves cancer-associated venous thromboembolism (vte) prediction :800--827

\bibitem{xie2024rna}
Xie C, Yang Y, Yu H, He Q, Yuan M, et~al. 2024.
Rna velocity prediction via neural ordinary differential equation.
\textit{Iscience} 27(4)

\bibitem{pinckaers2019neural}
Pinckaers H, Litjens G. 2019.
{Neural ordinary differential equations for semantic segmentation of individual colon glands}.
\textit{arXiv preprint arXiv:1910.10470}

\bibitem{yang2023neural}
Yang Z, Hu Z, Ji H, Lafata K, Vaios E, et~al. 2023.
A neural ordinary differential equation model for visualizing deep neural network behaviors in multi-parametric {MRI-based} glioma segmentation.
\textit{Medical physics} 50(8):4825--4838

\bibitem{ru2023attention}
Ru J, Lu B, Chen B, Shi J, Chen G, et~al. 2023.
Attention guided neural ode network for breast tumor segmentation in medical images.
\textit{Computers in Biology and Medicine} 159:106884

\bibitem{chen2020mri}
Chen EZ, Chen T, Sun S. 2020.
\textit{Mri image reconstruction via learning optimization using neural odes}.
In \textit{Medical Image Computing and Computer Assisted Intervention--{MICCAI} 2020: 23rd International Conference, Lima, Peru, October 4--8, 2020, Proceedings, Part II 23}, pp.  83--93. Springer

\bibitem{xu2021multi}
Xu J, Chen EZ, Chen X, Chen T, Sun S. 2021.
\textit{Multi-scale neural odes for 3d medical image registration}.
In \textit{Medical Image Computing and Computer Assisted Intervention--{MICCAI} 2021: 24th International Conference, Strasbourg, France, September 27--October 1, 2021, Proceedings, Part IV 24}, pp.  213--223. Springer

\bibitem{wu2024neural}
Wu Y, Dong M, Jena R, Qin C, Gee JC. 2024.
Neural ordinary differential equation based sequential image registration for dynamic characterization.
\textit{arXiv preprint arXiv:2404.02106}

\bibitem{qiu2023gram}
Qiu X, Shi S, Tan X, Qu C, Fang Z, et~al. 2023.
\textit{{Gram-based} attentive neural ordinary differential equations network for video nystagmography classification}.
In \textit{Proceedings of the IEEE/CVF international conference on computer vision}, pp.  21339--21348

\bibitem{jeong2022deep}
Jeong S, Jung W, Sohn J, Suk HI. 2022.
\textit{Deep geometrical learning for Alzheimer’s disease progression modeling}.
In \textit{2022 IEEE International Conference on Data Mining (ICDM)}, pp.  211--220. IEEE

\bibitem{hao2020learning}
Hao W, Vogt NM, Meng Z, Hwang SJ, Koscik RL, et~al. 2020.
\textit{Learning Amyloid Pathology Progression from Longitudinal PIB-PET Images in Preclinical Alzheimer's Disease}.
In \textit{2020 IEEE 17th International Symposium on Biomedical Imaging (ISBI)}, pp.  572--576. IEEE

\bibitem{cai2023osr}
Cai H, Dan T, Huang Z, Wu G. 2023.
\textit{Osr-net: Ordinary differential equation-based brain state recognition neural network}.
In \textit{2023 IEEE 20th International Symposium on Biomedical Imaging (ISBI)}, pp.  1--5. IEEE

\bibitem{fronk2025training}
Fronk C, Petzold L. 2025.
Training stiff neural ordinary differential equations with explicit exponential integration methods.
\textit{Chaos: An Interdisciplinary Journal of Nonlinear Science} 35(3)

\bibitem{kidger2021efficient}
Kidger P, Foster J, Li XC, Lyons T. 2021.
\textit{Efficient and Accurate Gradients for Neural SDEs}.
In \textit{Advances in Neural Information Processing Systems}, ed. M~Ranzato, A~Beygelzimer, Y~Dauphin, P~Liang, JW~Vaughan, pp.  18747--18761, vol.~34, pp.  18747--18761. Curran Associates, Inc.

\bibitem{giampiccolo2024robust}
Giampiccolo S, Reali F, Fochesato A, Iacca G, Marchetti L. 2024.
Robust parameter estimation and identifiability analysis with hybrid neural ordinary differential equations in computational biology.
\textit{NPJ Systems Biology and Applications} 10(1):139

\bibitem{kumar2025physics}
Kumar T, Kumar A, Pal P. 2025.
A physics-constrained neural ordinary differential equations approach for robust learning of stiff chemical kinetics.
\textit{Combustion Theory and Modelling} :1--16

\bibitem{rubanova2019latent}
Rubanova Y, Chen RT, Duvenaud DK. 2019.
Latent ordinary differential equations for irregularly-sampled time series.
\textit{Advances in neural information processing systems} 32

\bibitem{dupont2019augmented}
Dupont E, Doucet A, Teh YW. 2019.
Augmented neural odes.
\textit{Advances in neural information processing systems} 32

\bibitem{kanode}
Koenig BC, Kim S, Deng S. 2024.
{KAN-ODEs}: Kolmogorov–arnold network ordinary differential equations for learning dynamical systems and hidden physics.
\textit{Computer Methods in Applied Mechanics and Engineering} 432:117397

\bibitem{lu2019deeponet}
Lu L, Jin P, Karniadakis GE. 2019.
{DeepOnet}: Learning nonlinear operators for identifying differential equations based on the universal approximation theorem of operators.
\textit{arXiv preprint arXiv:1910.03193}

\bibitem{li2020neural}
Li Z, Kovachki N, Azizzadenesheli K, Liu B, Bhattacharya K, et~al. 2020{\natexlab{a}}.
Neural operator: Graph kernel network for partial differential equations.
\textit{arXiv preprint arXiv:2003.03485}

\bibitem{li2020fourier}
Li Z, Kovachki N, Azizzadenesheli K, Liu B, Bhattacharya K, et~al. 2020{\natexlab{b}}.
Fourier neural operator for parametric partial differential equations.
\textit{arXiv preprint arXiv:2010.08895}

\bibitem{tripura2023wavelet}
Tripura T, Awasthi A, Roy S, Chakraborty S. 2023.
A wavelet neural operator based elastography for localization and quantification of tumors.
\textit{Computer Methods and Programs in Biomedicine} 232:107436

\bibitem{cao2024laplace}
Cao Q, Goswami S, Karniadakis GE. 2024.
Laplace neural operator for solving differential equations.
\textit{Nature Machine Intelligence} 6(6):631--640

\bibitem{wang2021learning}
Wang S, Wang H, Perdikaris P. 2021{\natexlab{b}}.
{Learning the solution operator of parametric partial differential equations with physics-informed DeepONets}.
\textit{Science advances} 7(40):eabi8605

\bibitem{goswami2023physics}
Goswami S, Bora A, Yu Y, Karniadakis GE. 2023{\natexlab{b}}.
Physics-informed deep neural operator networks.
In \textit{Machine learning in modeling and simulation: methods and applications}. Springer

\bibitem{jafarzadeh2024peridynamic}
Jafarzadeh S, Silling S, Liu N, Zhang Z, Yu Y. 2024.
Peridynamic neural operators: A data-driven nonlocal constitutive model for complex material responses.
\textit{Computer Methods in Applied Mechanics and Engineering} 425:116914

\bibitem{jafarzadeh2025heterogeneous}
Jafarzadeh S, Silling S, Zhang L, Ross C, Lee CH, et~al. 2025.
Heterogeneous peridynamic neural operators: Discover biotissue constitutive law and microstructure from digital image correlation measurements.
\textit{Foundations of Data Science} 7(1):226--270

\bibitem{you2022physics}
You H, Zhang Q, Ross CJ, Lee CH, Hsu MC, Yu Y. 2022.
A physics-guided neural operator learning approach to model biological tissues from digital image correlation measurements.
\textit{Journal of Biomechanical Engineering} 144(12):121012

\bibitem{zhu2024neural}
Zhu Y, Peng B. 2024.
\textit{Neural Operator-Based Framework for Time Efficient Denoising of Displacement Fields in Ultrasound Elastography}.
In \textit{2024 IEEE International Conference on Systems, Man, and Cybernetics (SMC)}, pp.  3318--3323. IEEE

\bibitem{hong2024estimating}
Hong J, Min C, Lee B, Persad AR, Jeong JH, Park YH. 2024.
\textit{Estimating Aortic Pressure Waveform in a 1D Hemodynamic Model of the Human Arterial System using {DeepONet}}.
In \textit{2024 46th Annual International Conference of the IEEE Engineering in Medicine and Biology Society (EMBC)}, pp.  1--5. IEEE

\bibitem{kumar2024convolutional}
Kumar A, Zhi X, Ahmad Z, Yin M, Manbachi A. 2024.
Convolutional deep operator networks for learning nonlinear focused ultrasound wave propagation in heterogeneous spinal cord anatomy.
\textit{arXiv preprint arXiv:2412.16118}

\bibitem{wang2025ultrasound}
Wang J, Ostras O, Sode M, Tolooshams B, Li Z, et~al. 2025.
Ultrasound lung aeration map via physics-aware neural operators.
\textit{ArXiv} :arXiv--2501

\bibitem{drozdov2024fnoreg}
Drozdov NA, Sorokin DV. 2024.
\textit{{FNOReg}: Resolution-Robust Medical Image Registration Method Based on Fourier Neural Operator}.
In \textit{International Conference on Pattern Recognition}, pp.  163--177. Springer

\bibitem{zhang2022g2varphinet}
Zhang E, Spronck B, Humphrey JD, Karniadakis GE. 2022.
G2$\phi$net: Relating genotype and biomechanical phenotype of tissues with deep learning.
\textit{PLOS Computational Biology} 18(10):e1010660

\bibitem{husanovic2024deep}
Husanovic S, Egberts G, Heinlein A, Vermolen F. 2024.
Deep operator network models for predicting post-burn contraction.
\textit{arXiv preprint arXiv:2411.14555}

\bibitem{yin2022simulating}
Yin M, Ban E, Rego BV, Zhang E, Cavinato C, et~al. 2022.
Simulating progressive intramural damage leading to aortic dissection using {DeepONet}: an operator--regression neural network.
\textit{Journal of the Royal Society Interface} 19(187):20210670

\bibitem{guan2023fourier}
Guan S, Hsu KT, Chitnis PV. 2023.
Fourier neural operator network for fast photoacoustic wave simulations.
\textit{Algorithms} 16(2):124

\bibitem{chumburidze2025efficient}
Chumburidze M, Niminet V. 2025.
Efficient modelisation for complex geometries of tumor growth via thermo-elastic diffusion partial differential equations and artificial neural networks.
\textit{BRAIN. Broad Research in Artificial Intelligence and Neuroscience} 16(1):315--323

\bibitem{salvador2024liquid}
Salvador M, Marsden AL. 2024.
Liquid fourier latent dynamics networks for fast gpu-based numerical simulations in computational cardiology.
\textit{arXiv preprint arXiv:2408.09818}

\bibitem{hu2024state}
Hu Z, Daryakenari NA, Shen Q, Kawaguchi K, Karniadakis GE. 2024.
State-space models are accurate and efficient neural operators for dynamical systems.
\textit{arXiv preprint arXiv:2409.03231}

\bibitem{d2020bayesian}
Dandekar R, Chung K, Dixit V, Tarek M, Garcia-Valadez A, et~al. 2020.
Bayesian neural ordinary differential equations.
\textit{arXiv preprint arXiv:2012.07244} \url{https://arxiv.org/abs/2012.07244}

\bibitem{zouuq}
Zou Z, Meng X, Psaros AF, Karniadakis GE. 2024.
Neuraluq: A comprehensive library for uncertainty quantification in neural differential equations and operators.
\textit{SIAM Review} 66(1):161--190

\bibitem{meng2020composite}
Meng X, Karniadakis GE. 2020.
A composite neural network that learns from multi-fidelity data: Application to function approximation and inverse pde problems.
\textit{Journal of Computational Physics} 401:109020

\bibitem{lu2022multifidelity}
Lu L, Pestourie R, Johnson SG, Romano G. 2022.
Multifidelity deep neural operators for efficient learning of partial differential equations with application to fast inverse design of nanoscale heat transport.
\textit{Physical Review Research} 4(2):023210

\bibitem{song2013optimal}
Song Q, Bai J, Han D, Bhatia S, Sun W, et~al. 2013.
Optimal co-segmentation of tumor in pet-ct images with context information.
\textit{IEEE transactions on medical imaging} 32(9):1685--1697

\bibitem{tang2023explainable}
Tang X, Zhang J, He Y, Zhang X, Lin Z, et~al. 2023.
Explainable multi-task learning for multi-modality biological data analysis.
\textit{Nature communications} 14(1):2546

\bibitem{swanson2024virtual}
Swanson K, Wu W, Bulaong NL, Pak JE, Zou J. 2024.
The virtual lab: {AI} agents design new sars-cov-2 nanobodies with experimental validation.
\textit{bioRxiv} :2024--11

\bibitem{essam2025robin}
Essam~Ghareeb A, Chang B, Mitchener L, Yiu A, Szostkiewicz CJ, et~al. 2025.
Robin: A multi-agent system for automating scientific discovery.
\textit{arXiv e-prints} :arXiv--2505

\end{thebibliography}

\end{document}